\pdfoutput=1
\documentclass{article}
\usepackage{amsmath, amssymb, amsfonts}
\usepackage{bm} %
\usepackage{microtype}
\usepackage{enumitem}
\microtypesetup{tracking=false}
\usepackage{amsmath} %
\usepackage{amssymb}  %
\usepackage{amsfonts}
\usepackage{graphicx}
\usepackage[linesnumbered,ruled,vlined]{algorithm2e} 
\usepackage{epsfig} %
\usepackage{xspace}
\usepackage{url}
\usepackage{comment}
\usepackage[dvipsnames,table]{xcolor}
\usepackage{colortbl}          %
\usepackage[font=footnotesize]{caption}
\usepackage{booktabs} %
\usepackage{tikz}
\usetikzlibrary{positioning,shapes.geometric,arrows,fit,shapes.symbols,svg.path,tikzmark,calc}
\usepackage{textcomp}
\usepackage{listings}
\usepackage{subcaption}
\pgfdeclarelayer{background}
\pgfdeclarelayer{foreground}
\pgfsetlayers{background,main,foreground}
\usepackage{adjustbox}
\usepackage{array}
\usepackage{multirow}
\usepackage{dsfont}

\usepackage{siunitx}
\usepackage{mathtools}
\usepackage{enumitem}
\usepackage{physics}

\usepackage{amsthm}

\usepackage{adjustbox} %
\usepackage{subcaption}
\usepackage{wrapfig}

\usepackage{multirow, makecell} %

\usepackage[dvipsnames,table]{xcolor}         %
\usepackage{titletoc}        %
\usepackage{hyperref}       %
\hypersetup{
    colorlinks=true,     
    urlcolor= WildStrawberry,
    citecolor=MidnightBlue,
    }

\renewcommand{\arraystretch}{1.3}

\definecolor{pregrasp}{RGB}{255,225,235}     %
\definecolor{contact}{RGB}{255,245,220}      %
\definecolor{grasp}{RGB}{230,250,230}        %
\definecolor{release}{RGB}{220,235,255}      %
\definecolor{postgrasp}{RGB}{235,220,245}     %

\SetKwFor{ParallelFor}{for}{do in parallel}{endfor}
\SetKwInOut{Input}{Input}
\SetKwInOut{Output}{Output}

\usepackage{xspace}

\usepackage{listings}
\usepackage{xcolor}

\usepackage{lmodern}
\usepackage[most]{tcolorbox}
\usepackage{microtype}

\hypersetup{
  colorlinks=true,
  linkcolor=blue!50!black,
  urlcolor=blue!50!black,
  citecolor=blue!50!black
}

\definecolor{SkillBack}{RGB}{248,249,251}
\definecolor{SkillFrame}{RGB}{210,216,224}
\definecolor{SkillTitleBack}{RGB}{236,240,246}

\tcbset{
  skillbox/.style={
    enhanced,
    breakable,
    colback=SkillBack,
    colframe=SkillFrame,
    colbacktitle=SkillTitleBack,
    coltitle=black,
    boxrule=0.45pt,
    titlerule=0pt,
    arc=2pt,
    outer arc=2pt,
    left=6pt,
    right=6pt,
    top=5pt,
    bottom=5pt,
    before skip=6pt,
    after skip=6pt,
    fonttitle=\bfseries,
    attach boxed title to top left={xshift=5pt,yshift*=-1.5mm},
    boxed title style={
      colback=SkillTitleBack,
      colframe=SkillFrame,
      boxrule=0.35pt,
      arc=2pt,
      left=4pt,
      right=4pt,
      top=1pt,
      bottom=1pt
    }
  }
}

\newcommand{\skillcard}[4]{%
\begin{tcolorbox}[skillbox,title={\texttt{#1}}]
\small
\noindent\textbf{Description.} #4\par\vspace{2pt}
\noindent\textbf{Inputs.} #2\par\vspace{2pt}
\noindent\textbf{Outputs.} #3
\end{tcolorbox}%
}

\lstdefinelanguage{Markdown}{
    keywords={},
    sensitive=false,
    morecomment=[l]{\#},
}
\usepackage{wrapfig}

\definecolor{spatclr}{RGB}{215,226,244}   %
\definecolor{goalclr}{RGB}{208,235,200}   %
\definecolor{intclr} {RGB}{230,222,247}   %

\usepackage{tabularx}
\usepackage{pifont} %
\usepackage[most]{tcolorbox}
\tcbset{colback=gray!10, colframe=gray!80!black, boxrule=0.5pt, arc=1mm, left=1mm, right=1mm, top=1mm, bottom=1mm, boxsep=1mm}

\usepackage[final]{corl_2026} %

\title{GaP: A Graph-as-Policy Multi-Agent Self-Learning Harness For Variational Automation (VA) Tasks}

\author{
\normalfont
    Kaiyuan Chen$^{3,*}$ \quad
    Shuangyu Xie$^{3,*}$ \quad
    Letian Fu$^{3}$ \quad
    Justin Yu$^{3}$ \quad
    William Pacini$^{3}$ \\
    Sandeep Bajamahal$^{3}$ \quad
    Hudson Kim$^{3}$ \quad
    Jaimyn Drake$^{3}$ \quad
    Daehwa Kim$^{3}$ \quad
    Haoru Xue$^{3}$ \\
    Jonathan Francis$^{1}$ \quad
    Christian Juette$^{1}$ \quad
    Peter Schaldenbrand$^{1}$ \quad
    Muhammet Yunus Seker$^{1}$ \\
    Ruwan Wickramarachchi$^{1}$ \quad
    Uksang Yoo$^{3}$ \quad
    Guanzhi Wang$^{2}$ \quad \\
    Adithyavairavan Murali$^{2}$ \quad
    Balakumar Sundaralingam$^{2}$ \\
    S. Shankar Sastry$^{3}$ \quad
    Spencer Huang$^{2}$ \quad
    Yuke Zhu$^{2}$ \quad
    Linxi ``Jim'' Fan$^{2}$ \quad
    Ken Goldberg$^{3}$ \\[0.5em]
    $^{1}$ Bosch \qquad
    $^{2}$ NVIDIA \qquad
    $^{3}$ University of California, Berkeley \\[0.25em]
    $^{*}$ equal contribution
}
\usepackage{algorithm}
\usepackage{algorithmic}
\usepackage{booktabs}
\usepackage{graphicx} %
\begin{document}
\maketitle

\vspace{-20pt}
\begin{center}
    \centering
    
    \captionsetup{type=figure}
    \includegraphics[width=\linewidth]{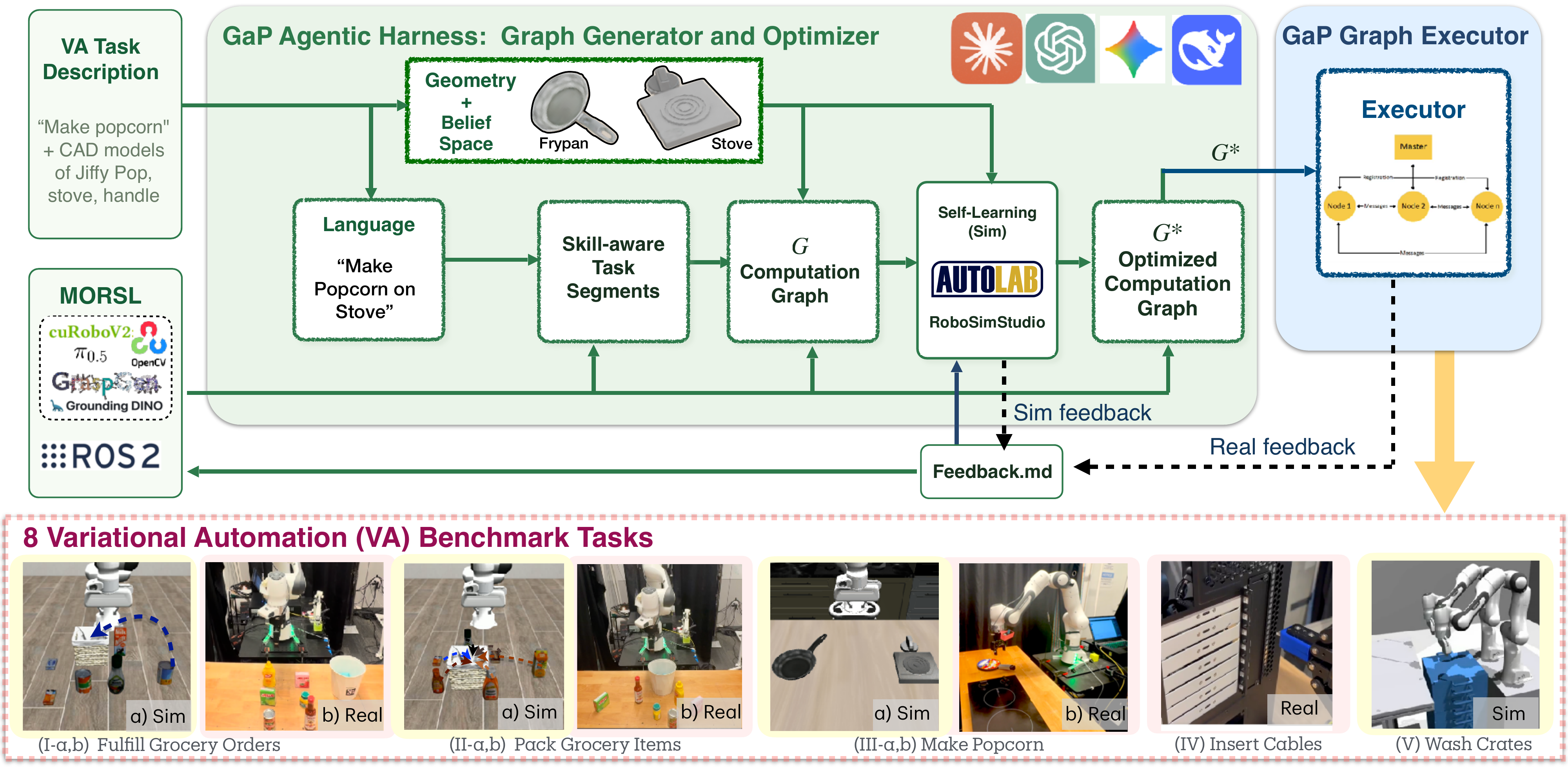}
    \captionof{figure}{\textbf{GaP} system architecture. Given a  {``Variational Automation (VA)'' }task specification, GaP uses a multi-agent harness for coding tools such as Claude~\cite{anthropic2024claude35} and Gemini~\cite{google2025gemini25} to automatically generate computation graphs that include ``skill'' nodes from a Modular Open Robot Skill Library (MORSL), which includes model-based procedures (e.g. ROS~\cite{macenski2022robot}) and model-free policies (e.g. GraspGen~\cite{murali2025graspgen}). GaP then orchestrates self-learning using simulation (e.g. RoboSimStudio~\cite{chen2026robosimstudio}) to iteratively refine the graph, which can then be interpreted without the agents on an edge device for persistent execution over time. Bottom: 8 Variational Automation Benchmark Tasks (4 Sim and 4 Real).}
    \label{fig:scene-title}
\end{center}%

\begin{abstract}
    For robots to work reliably in commercial and industrial applications, can recent advances in agentic coding systems combine interpretable robot programming with the open-world adaptability of model-free policies?  We focus on ``Variational Automation'' (VA), a class of tasks that have larger variations in object geometry and pose than fixed automation.  Model-free policies often struggle to close the reliability gap for VA tasks, which must be executed persistently and reliably in commercial and industrial applications. Motivated by prior work on Task and Motion Planning (TAMP) and the Robot Operating System (ROS), we introduce Graph-as-Policy (GaP), a multi-agent coding harness that generates directed computation graphs with perception, planning, and control nodes from a Modular Open Robot Skill Library (MORSL).
    GaP then generates an internal simulation environment to rehearse task instances with different graphs in parallel to iteratively refine the graph structure and parameters to improve success rates and throughput.   Evaluation with 8 new open VA task benchmarks, 4 in simulation and 4 in real-world, suggests that GaP can achieve success rates that significantly outperform baselines. Details, code, and data will be posted online: \url{https://graph-robots.github.io/gap}
\end{abstract}


\section{Introduction}
A majority of robot learning research focuses on generalist robotics, where a robot must perform a broad variety of tasks.  Most of this research has focused on model-free end-to-end Vision-Language-Action (VLA) models ~\cite{kim24openvla, pi0_2024, open_x_embodiment_rt_x_2023, fang2026molmoact2, black2025pi_, zawalski2025robotic}. However, fully generalist robots struggle to achieve commercial / industrial levels of reliability \cite{gao2025taxonomy} and there is increasing interest in how robot learning could be useful for more specialized task classes~\cite{goldberg2026generalists}. In this paper, we identify a class of ``Variational Automation (VA)'' tasks that differ from ``fixed automation'' (which blindly repeats the same motions, e.g., spot welding or spray painting).  In VA tasks, a robot persistently performs varying instances of a task with non-trivial variation in the geometry and pose of objects (e.g., to sort packages, make coffee in a cafe, or build sandwiches in a commercial kitchen). Today, fixed automation in logistics, service, agriculture, and manufacturing \cite{solowjow_indust_2020, adebola2024automating, xie_energy_2025, adebolagarden2023} is set up and tuned by humans using rigorous, classical engineering methods. This provides high reliability and throughput, and human effort can be justified when amortized over years of repetitive performance. However, even more human effort is required to set up and tune VA systems for reliable performance.

Recent advances in Large Language Models (LLM)~\cite{team2026qwen3, yang2025qwen3,guo2025deepseek} and Vision-Language Models (VLM)~\cite{radford2021learning, bordes2024introduction, geminiroboticsteam2025geminiroboticsbringingai, huang2023voxposer,ning2025prompting} are rapidly improving agentic coding~\cite{claudecode}, semantic reasoning~\cite{chenrobo2vlm} and zero-shot generalization~\cite{ji2025robobrain}. Agentic coding for robotics has the potential to integrate model-based and model-free paradigms to rapidly generate interpretable and reliable robot control systems. Agentic coding is new and advancing rapidly, but remains prone to hallucinations and constraint violations. Thus, a primary challenge is to develop effective methods for ``steering'' coding agents to produce desired outputs.  Coding agents are controlled with a ``harness'' -- a body of text provided as a prompt to the agents that describe available software resources, constraints, and performance metrics~\cite{anthropicharness}.  The harness also provides interfaces that allow the LLM to compile, execute, and manage generate-execute-observe-evaluate loops.

The use of LLM coding agents for robotics was explored as early as 2022 with a series of papers on 
Code-as-Policy (CaP) ~\cite{codeaspolicies2022, YinG-RSS-25}.
General coding agents have improved dramatically since then by training on increasingly large coding datasets.  The recent CaP-X paper provides an open suite of CaP benchmarks, agents, and encouraging results using coding agents that were state-of-the-art as of January 2026 ~\cite{fu2026cap}.
These single-agent CaP approaches are somewhat unstructured, prompting the coding agent to generate free-form python code.  For more complex tasks, the context window can grow sufficiently large so that it is difficult for the agent to obey constraints and converge on reliable code.  Individual agents are also prone to hallucinations and ``cheating'', where they make up nonexistent skills or create trivial success metrics to artificially ''solve'' robot coding tasks.  These challenges are exacerbated with \emph{multi-agent} coding systems, where multiple agents operate in parallel.

To address these issues, we propose Graph-as-Policy (GaP), a harness structure that encapsulates modular robot functions into independent ``nodes'' in a directed graph that can be managed by a hierarchical multi-agent system where optimizing each node can be assigned to a specific agent.  This structure also helps limit the size of each agent's context window and separates the graph generation from the graph testing to reduce incentives for individual coding agents to ``cheat'' in order to achieve requested objectives. GaP is inspired in part by the architecture of robot Task and Motion Planning (TAMP) systems~\cite{kaelbling2011hierarchical,garrett2021integrated, shen2026tiptop} which use graphical hierarchies to ensure safety and the Robot Operating System (ROS)\cite{macenski2022robot} which is based on a computation graph structure.  Graphical models support modularity, reuse, and composability of functions in non-robotic and robotic applications.   In GaP, a robot policy is a computation graph composed of atomic ``skill'' nodes, such as retrieving a camera frame, running inference on a perception model, or planning and executing a motion trajectory. 

Given a VA task description, GaP starts with an Orchestration Agent that partitions a VA task into functional segments and instructs appropriate Skill Agents to synthesize localized, functional subgraphs of atomic nodes for its assigned segment. The orchestrator then aggregates and wires these subgraphs together into an executable computation graph.  GaP then initiates a multi-agent self-learning harness that coordinates LLM and VLM agents to autonomously generate, simulate, evaluate, robot computation graphs in a loop with sampled task instances to increase a weighted combination of success rate and throughput until the combination reaches a plateau.  The resulting computation graph is then sent to an edge-based graph interpreter for repeated execution on the robot.

To evaluate GaP, we present 8 new open VA benchmarks. The first 6 -- based loosely on persistent commercial applications -- are (I-a,b) Fulfill Grocery Orders, (II-a,b) Pack Grocery Items, and (III-a,b) Make Popcorn, where `a' is in sim and `b' is in real. These 6 ongoing tasks use common kitchen items and one Franka robot arm from the LIBERO~\cite{liu2023libero} benchmark. The other two VA benchmarks -- based loosely on datacenter and industrial applications -- are (IV) Insert USB-C Cables, and (V) Wash Crates. Results suggest that GaP can achieve success rates that significantly outperform baselines, including $\pi_{0.5}$~\cite{black2025pi_} and MolmoAct2~\cite{fang2026molmoact2}.

This paper makes the following contributions:
(1) the VA task class for robot learning with 8 open VA benchmarks; (2)  a computation graph structure for agentic robotics; %
(3) GaP, an implemented multi-agent harness that autonomously decomposes natural language specifications to collaboratively generate robot computation graphs;
(4) an open, evolving Modular Open Robot Skill Library (MORSL) with 51 initial skills;
(5) a self-learning approach to graph rehearsal and optimization using internal simulations built with RoboSimStudio~\cite{chen2026robosimstudio}; and
(6) Experimental data comparing GaP with TAMP, model-free, and other baselines and simulations on the first 6 VA benchmarks, and performance results from GaP on the other 2 VA benchmarks.

\section{Related Work}

\textbf{Variational Automation}.
The term ``automation'' is commonly used to describe robot systems that persistently perform repetitive tasks over hours, weeks, months, or years for logistics, manufacturing, healthcare, service, agriculture, and other applications. To be successful, commercial and industrial automation must achieve desired throughput (units per hour: success rate divided by cycle time) at a cost that will provide a desired return on investment. Accordingly, research in automation extends research in robotics by focusing on reliability, cost, safety, ease of use, and other factors required for successful production deployment.

In this paper, we propose ``Variational Automation (VA)'' as a class of tasks that differ from Generalist Robotics (GR) and Fixed Automation (FA). In FA, a robot persistently performs identical instances of a task with objects of identical geometry. In box stacking, for example, FA would consist of moving identical boxes from a common initial pose on a conveyor belt into a fixed pallet arrangement. In FA, variations in the environment and object shape and pose are minimal. 

For GR, a robot performs a variety of different tasks (e.g., placing groceries into a refrigerator, or 
domestic cleaning, folding, kitchen pick-and-place) in different homes, where environments vary considerably and objects have variable geometry and highly variable initial poses. Recent robot learning research predominantly targets highly unstructured Generalist Robotics (GR), such as household robots executing diverse chores across novel environments using a single generalist model-free VLA policies~\cite{levine2016end, chi2025diffusion, kim24openvla, pi0_2024, black2025pi_}.

In VA, by contrast, a robot persistently performs varying instances of the task: the boxes have variable geometry (different SKUs), arrive in a distribution of varying initial poses, and must be densely packed into varying pallet arrangements. As most robot learning research focuses on GR, and robot learning may not be required for FA, we focus on robot learning for VA, aiming to reduce the human effort required to set up VA systems.

\textbf{Modular Robot Control Methods}.
Early robotic systems programmed policies using classical modular components. The Shakey robot~\cite{nilsson1998artificial} in the late 1960s
decomposed its architecture into 3 functional components:
sensing, planning, and execution~\cite{nilsson2014principles}. System architectures like the Robot Operating System (ROS) \cite{macenski2022robot} utilize explicit directed graphs to route data, manage dependencies, and ensure reliable execution.
Concurrently, Task and Motion Planning (TAMP)~\cite{kaelbling2011hierarchical,garrett2021integrated} approaches jointly solve discrete task planning and continuous motion planning problems, enabling the satisfaction of constraints involving both high-level action sequencing and low-level geometric feasibility. Behavior Trees (BTs)~\cite{colledanchise2018behavior} encode reactive task logic and failure recovery, and recent work shows LLMs can generate BTs with high success rates~\cite{izzo2024btgenbot}. GaP combines the typed dataflow topology of the ROS computation graph with the reactive task logic and failure recovery of BTs, enabling (a) static type checking and verification, (b) explicit failure recovery loops, and (c) simulation-guided graph optimization.

Code-as-Policy (CaP) uses agents to write robot control code, suggesting an alternative to manual coding of model-based methods and pure end-to-end learning of model-free policies~\cite{codeaspolicies2022}. 
Extensions such as ALGARA \cite{coteagentic}, CaP-X \cite{fu2026cap}, CodeDiffuser~\cite{yin2025codediffuser}, GRAPPA~\cite{bucker2026grappa}, and Maestro \cite{shi2025maestro} use more recent LLMs to generate robot code. Similarly, systems like TiPToP~\cite{shen2026tiptop} incorporate LLMs with classical robot TAMP~\cite{wang2024llm, curtis2025trust, yang2025guiding, kumar2026open}; TAMP can achieve general open-vocabulary manipulation and transfer to new tasks with minimal tuning, and TiPToP integrates Gemini perception, M2T2 grasp planning, and cuTAMP. GaP explores an alternative modular architecture that specializes a persistent graph to a VA task class.

Code-as-Policy (CaP)
is an alternative to manual coding of model-based methods and pure end-to-end learning of model-free policies \cite{codeaspolicies2022}, ALGARA \cite{coteagentic}, and CodeDiffuser~\cite{yin2025codediffuser} utilized LLMs for open-vocabulary script generation to generate robot code.
GaP tempers the flexibility of CaP approaches by introducing a graph structure, allowing GaP to harness the open-world adaptivity of pretrained LLM coding agents while maintaining a structured, interpretable graph structure to support persistent Variational Automation.  

\textbf{Self-Improving Agentic Workflows For Robotic Control}.
Outside of robotics, Large Language Models (LLMs) have demonstrated exceptional capabilities in dynamic coding, complex software integration, and API orchestration \cite{wu2024introducing, team2026qwen3, hou2025model, openai2024gpt4o, anthropic2024claude35, google2025gemini25}. By embedding these capabilities into structured agentic workflows, LLMs can autonomously synthesize solutions for open-ended tasks~\cite{zhang2025aflow}. A foundational example is Voyager \cite{wang2023voyager}, which uses LLMs to iteratively write, refine, and execute Minecraft game playing skills to continually expand a curated library of reusable skills. In these frameworks, a ``skill'' is typically defined as a discrete function with strict semantic contracts. Recent literature has begun to explore the iterative optimization of agentic systems through language-based failure reasoning \cite{zehle2026promptolution} and multi-agent prompt optimization frameworks \cite{agrawal2025gepa, lee2025feedback}.

To improve code generation quality and performance, CaP-X~\cite{fu2026cap} used a VLM to provide feedback before and after execution (i.e. Visual Differencing), but the VLM can suffer from hallucinations and cannot handle geometric and numerical information such as motion feasibility. 
Building on the structure of Blox-Net~\cite{goldberg2025bloxnet}, which uses physical experiments to improve LLM-generated task plans, GaP uses multiple LLM agents to generate a robot computation graph and iteratively improve it using simulation experiments.

\section{Problem Formulation}

\paragraph{Assumptions}
For Variational Automation tasks that will be repeated over extended periods, we assume the workcell environment, robot, and sensors are known and fixed. Furthermore, we assume the range of potential objects and the range of initial object poses are known. These assumptions are part of the VA setting rather than oracle information: unlike generalist robotics, VA tasks are defined by a known workcell and bounded operating envelope, allowing systems to use object models, calibrated sensors, and reusable automation skills when available. Because object meshes and pose ranges are part of the task input, the internal simulation is instantiated directly from them without scene reconstruction, system identification, or domain randomization.

\paragraph{Variational Automation Task Class}
We formalize a Variational Automation (VA) Task using the tuple $\mathcal{T} = \langle \mathcal{L}, \mathcal{E}, \mathcal{R}, \mathcal{O}, \mathcal{X}, \mathcal{B}, \mathcal{J} \rangle$:

\begin{itemize} [nosep, leftmargin=*]
    \item $\mathcal{L}$, Language Space: Natural language instructions and semantic descriptors that describe desired task behavior, providing high-level context for decomposing the task into segments.
    \item $\mathcal{E}$, is a known stationary workspace environment (e.g., workcell) that establishes world frame $\mathcal{W}$ and occupancy map $\mathcal{M}_E$, providing the support surfaces that constrain the feasible pose space of all entities to the non-colliding subset of $SE(3)$.
    \item $\mathcal{R}$, is the {robot and sensor configuration}, specifying the robot URDF, joint limits, gripper configuration, camera spec, and camera placement.
    \item $\mathcal{O}$, Object Set: The set of all possible objects in the task, comprising both rigid objects $\{o_i\}$ with 3D models $\mathcal{M}_i$ and articulated entities $\{\kappa_j\}$ (e.g., drawers, knobs) with defined kinematic joint limits $[\theta_{\min}, \theta_{\max}]$.
    \item {$\mathcal{X}$, State Space:}  The product space of the robot joint configurations, the $SE(3)$ poses of all rigid objects, and the kinematic states of articulated entities: $\mathcal{X} = \mathcal{C}_{robot} \times SE(3)^n \times \mathbb{R}^m$.
    \item {$\mathcal{B}$, Belief Space:} Describes the distribution of objects and poses in instances of the task. A specific instance is sampled $\mathbf{x}_i \sim p(\mathbf{x} \mid\mathcal{X})$, introducing variations such as: (i) \textit{Structured Priors:} Position sampled uniformly over a volume $\mathcal{V}$ ($\mathbf{x}_i \sim \text{Uniform}(\mathcal{V})$) and orientation ranges, and (ii) \textit{Empirical Distributions:} Multi-modal distributions estimated from real-world demonstrations or perception (e.g., point cloud registration).
    \item $\mathcal{J}$, Reward function: A multi-objective reward function used to evaluate success and efficiency, defined as:$\mathcal{J} = w_s \cdot \mathbb{I}(\text{success}) + w_t \cdot \Phi$ where $\mathbb{I}(\cdot)$ is the success indicator, $\Phi$ is the throughput (success rate / cycle time), and $w_s, w_t$ are weighting factors.

\end{itemize}

Given a task $\mathcal{T}$, a task instance $\tau_i = \langle o_i, x_i \rangle$ is drawn from $\mathcal{O}$ and $\mathcal{B}$ respectively.
   
\paragraph{Policy Representation via Directed Execution Graphs}

Graph-as-policy (GaP) represents a robot policy $\pi(a \mid \mathbf{x}, \mathcal{T})$  as a directed computation graph $\mathcal{G} = (V, E)$ to complete all task instances   $\forall \tau_i = \langle o_i \subseteq \mathcal{O}, x_i \sim \mathcal{B} \rangle \in \mathcal{T}$. The graph $\mathcal{G}$ consists of modular, atomic functional units called \emph{nodes}, connected by edges that dictate both data flow and execution logic. The components are defined as follows: \textbf{Nodes ($V$):} Each node $n \in V$ represents a functional primitive for manipulation, perception, or a segment of executable code. Each node encapsulates a single, well-defined robotic operation with a typed input/output signature.
Nodes can be grouped into \textbf{skills},  a natural-language specification that instructs an LLM agent how to configure and compose a set of atomic nodes on the execution graph for a defined sub-task. 
\textbf{Edges ($E$):} A directed edge $e = (n_i, n_j) \in E$ represents a data or logic dependency. {Data edges} route a producer node's output to a consumer node's input (e.g., feeding the output of an object-centering node into a planar-surface-orienting node) and implicitly induce execution order through their dependencies; independent branches may execute concurrently. {Control edges} are condition branches and carry a predicate over node outputs.

\paragraph{Problem Definition}

Given a task class $\mathcal{T} = \langle \mathcal{L}, \mathcal{E}, \mathcal{R}, \mathcal{O}, \mathcal{X}, \mathcal{B}, \mathcal{J} \rangle$, GaP must synthesize a robust execution graph $\mathcal{G}^*$ that generalizes across the entire belief space $\mathcal{B}$. Formally, for any task instance $\tau_i \in \mathcal{T}$ and given a real-time scene observation $\mathcal{I}$ (comprising multi-view image sets from static and wrist cameras), the graph executor interprets
the computation graph $\mathcal{G}^*$ by invoking its skill nodes and following its data and control edges.  This execution induces a closed-loop robot policy $\pi_G(a \mid \mathcal{I})$, which maps observations and graph state to robot actions in order to achieve the goal state specified by $\mathcal{L}$.  We define the optimized graph $\mathcal{G}^*$ as one that maximizes performance across all instances in the variational class:~$\mathcal{G}^* = \arg\max_{\mathcal{G}} \mathbb{E}_{x_i \sim \mathcal{B}} \left[ \mathcal{J}(\pi(a\mid\mathcal{I}, \mathcal{G})) \right]$. 

The optimized graph is then sent to an edge-based interpreter for repeated execution on the robot.

\section{Graph-as-Policy}
\label{sec:design}
The GaP architecture is illustrated in Figure \ref{fig:scene-title}.  As an example, consider the Make Popcorn VA task. Given a natural language specification and geometric object models, the multi-agent GaP harness first partitions the high-level objective into semantic segments such as ‘turn on the knob’, ‘pick up the popcorn pan’, etc. Next, drawing from the Modular Open Robot Skill Library (MORSL), GaP maps these segments into atomic robotic skills to synthesize an initial computation graph $\mathcal{G}$.  During self-learning, the computation graph is evaluated using an internal simulation to iteratively refine the execution graph topology and node parameters prior to deployment. The optimized robot computation graph is then sent to an external interpreter for repeated execution on the physical robot.

\subsection{Modular Open Robot Skill Library (MORSL)}
The MORSL library uses agentic tool-use conventions (e.g., Anthropic's
\texttt{Skill.md}~\cite{anthropicskill}) with extensions for graph declarations. Each skill declares its inputs, outputs, semantic parameters, and pre-conditions, so the agent can decide both when to invoke it and how to wire it into the graph. Examples of the 51 initial MORSL skills include perception (SAM2~\cite{ravi2024sam}/3~\cite{carion2025sam3segmentconcepts}, Grounding DINO~\cite{liu2023grounding}, OWL-ViT~\cite{minderer2022simple}, Molmo~\cite{deitke2024molmo}, general-purpose VLM~\cite{geminiroboticsteam2025geminiroboticsbringingai, kamath2025gemma}; 15 skills), grasp planning (Contact GraspNet~\cite{sundermeyer2021contact}, GraspGen~\cite{murali2025graspgen}, M2T2~\cite{yuan2023m2t2}; 5 skills), motion planning (cuRobo~\cite{sundaralingam2023curobo} and cuRobov2~\cite{sundaralingam2026curobov2}; 8 skills), 2D and 3D vision  utilities with NumPy and OpenCV (e.g., DBSCAN for point cloud processing; 15 skills), and 8 additional verification and control primitives including
(*) {Cartesian Linear Motion Planning with CuRobo}; 
(*) {Robot Operating System (ROS) Translator };  
(*) {Visuomotor Interactive Perception Policies}.  The full catalog can be found in the Appendix A.

\subsection{Self-Learning through Internal Simulation Rehearsal}
\label{sec:optimization}
\floatname{algorithm}{Self-Learning-Algorithm}
\begin{wrapfigure}{r}{0.55\textwidth}
\vspace{-24pt}
\begin{minipage}{0.55\textwidth}
\begin{algorithm}[H]
\scriptsize 
\DontPrintSemicolon
\caption{Rehearsal-based Graph Optimization}
\label{alg:graph_opt}
\SetKwInOut{KwInput}{In}
\SetKwInOut{KwOutput}{Out}
\noindent\textbf{Input:} $\mathcal{T} = \langle \mathcal{L}, \mathcal{E}, \mathcal{R}, \mathcal{O}, \mathcal{X}, \mathcal{B}, \mathcal{J} \rangle$,  Iterations $M$, Parallel Rollouts $N$
\noindent\textbf{Output:} Optimized Task Graph $\mathcal{G}^*$
\BlankLine
\textbf{Initialization:}\;
$\mathcal{G}_0 \leftarrow \mathrm{Graph\_Init}(\mathcal{L})$\;
\quad\tcp*[r]{\tiny Initial graph from language}
$\mathcal{S} \leftarrow \mathrm{Build\_Scene}(\mathcal{G}, \{\mathcal{G}_i\})$\;
\quad\tcp*[r]{\tiny Instantiate sim environment}
\For{$j \leftarrow 1$ \KwTo $M$}{
    \tcp{Step 1: Scene Variational Sampling}
    $\{\hat{s}_i\}_{i=1}^N \sim \mathcal{B}$\;
    \quad\tcp*[r]{\tiny Sample $N$ instances}
    \tcp{Step 2: Parallel Rehearsal}
    \ParallelFor{$i \leftarrow 1$ \KwTo $N$}{
        $\tau_i \leftarrow \mathrm{Rehearsal}(\hat{s}_i, \mathcal{G}_{j-1})$\;
        \quad\tcp*[r]{\tiny Rollout policy}
        $F_i \leftarrow \mathrm{Analyze_\_Failure}(\tau_i, \mathcal{G}, \{\mathcal{G}_i\})$\;
    
    }
    \tcp{Step 3: Graph Refinement}
    $\mathcal{G}_j \leftarrow \mathrm{Graph\_Update}(\{F_1, \dots, F_N\})$\;
    \quad\tcp*[r]{\tiny Optimize via LLM}
}
\Return $\mathcal{G}^* \leftarrow \mathcal{G}_M$\;
\end{algorithm}
\end{minipage}
\vspace{-8pt}
\end{wrapfigure}
To iteratively learn to improve a graph using task instances, we execute the computation graph, provide feedback, update the graph, and iterate. 
As described in self-learning Algorithm~\ref{alg:graph_opt}, GaP uses RoboSimStudio~\cite{chen2026robosimstudio}, an agentic toolkit for physics-based simulation of robot manipulation tasks, to instantiate the internal simulation that renders the visualization for the perception nodes, simulates physics, and computes contacts. 
To generate feedback, GaP registers the robot and object state before and after each rehearsing node, together with contact queries between the robot and objects, to compute differences and infer the motion outcome. 
The optimization process begins with task instance sampling, where the system generates $N$ parallel task instances  $\{{\tau}_i\}_{i=1}^N$. These task instances are instantiated by sampling from the belief space $\mathcal{B}$, which represents the probability distribution over potential object poses, initial states, and kinematic configurations within the workspace $\mathcal{E}$. 
GaP generates the initial graph $\mathcal{G}_0$ through multi-agent harness. 
Then, a parallel rehearsal starts across these $N$ scenes to evaluate graph performance. In instances where the rollout fails to meet success criteria, GaP analyzes the physical execution data to isolate geometric root causes within the occupancy or articulated entities. The graph is updated by the agents with Graph Update using a typed action set: (1) set a node's declared inputs, such as perception prompts; (2) rewire node inputs and outputs; (3) rewrite an inline Python script, such as format transformations and arithmetic; (4) swap skill nodes; and (5) verify the result with static type checking and simulation. Proposed updates that degrade success rate or cycle time on 100 fresh task instances are rejected, and the loop iterates until performance reaches a plateau or terminates with failure reasons.

\section{Experiments}

\subsection{Simulation and Real-World Variational Automation Benchmarks}
We present 8 VA benchmark tasks, including 4 in simulation and 4 in real-world.
Six of these are inspired by  LIBERO~\citep{liu2023libero}, the de facto standard for evaluating VLA policies in simulation.

\textbf{Benchmark (I-a,b): Fulfill Grocery Orders (Sim and Real)} Each instance of this task involves locating a designated target item and placing it into the basket. The original LIBERO~\cite{liu2023libero} benchmark did not vary the position of the  (\textit{objects}) which allows for overfitting to the demonstration examples. To address this, the {LIBERO-Pro}~\citep{zhou2025liberopro} evaluation suite swaps target and basket object positions at test time (\textit{object\_swap}). We introduce four extensions for our VA Benchmarks: 1) \textit{X-Y 20$\times$20 {\small{$cm^2$}}}, where the position of each object varies uniformly within a 20$\times$20 {\small{$cm^2$}} zone centered around the original LIBERO object pose; 2) \textit{basket\_swap}, which swaps the target item and basket location; 3) \textit{permutation}, which swaps target item and distractor item positions; and \textit{mixed\_all} including all three variation types. The physical benchmark directly mirrors this setup.
To avoid object collisions during task instance generation, we utilize Minkowski sum~\cite{minkowski1896geometrie} operations to ensure item singulation.  The real version of this benchmarks uses a Franka arm with wrist camera and  real items from a grocery store.  

\textbf{Benchmark (II-a,b): Pack Grocery Items (Sim and Real)} We modify Benchmark I-a,b to create multi-object pick-and-place tasks, as in a grocery checkout line, where the robot is assigned to pack 6 objects into a container without needing to identify or select any specific items. We quantify the success rate as the number of items (out of 6) placed into the basket after 6 grasping attempts. 

\textbf{Benchmark (III-a,b): Make Popcorn (Sim and Real)} We use the LIBERO frypan, stove, and knob assets to create a manipulation task that requires the robot to sequentially turn on a stove, pick up the pan handle, place it on the stove, remove the pan, and turn off the stove.  For the real version of this benchmark, we use the Franka Robot arm with wrist-camera, a portable stove from Amazon and Jiffy-Pop popcorn.

\textbf{Benchmark (IV): Insert USB-C Cables (Real)} This benchmark task requires a UR5 robot arm with the ZED Mini wrist camera to execute a series of USB-C cable insertions and extractions in a bank of 7 sockets.  The robot uses internal force torque feedback to probe the insertion when the target port goes beyond the camera's field of view.  The benchmark includes two varying conditions. First, we vary the pose of the cable ports, testing three distances in 5 {\small{$cm$}} increments and three angles in 15° increments. Second, we vary the text prompt to specify five distinct goals: targeting individual ports, inserting in ascending order, descending order, odd-indexed ports, and even-indexed ports.  

\textbf{Benchmark (V): Wash Crates (Sim) } This simulation benchmark simulates a real industrial crate-washing use case where two Franka robot arms must cooperatively grasp a crate by narrow slits on its sides, lift it off a stack, flip it, and place it onto a washing machine table.  Each instance varies the initial pose of the crate, varying yaw rotations of up to $\pm15^{\circ}$ and horizontal displacements of up to $\pm2.5$ {\small{$cm$}}.
The benchmark provides a language instruction \textit{``Cooperatively lift the top crate off the stack with both Franka arms, flip it over, and place it on the washing machine table,''}. (Fig.~\ref{fig:crate-wash-setup}).

\subsection{Baselines}
Each cell in Table 1 represents the outcome from trials with 100 task instances of Benchmarks I-a and II-a (in simulation) for a total of over 5000 simulation trials.  We run CaP-X~\cite{fu2026cap} which uses a single agent and is given only an image of the initial task instance and natural language instructions. We note that this is not a fair comparison as GaP is designed for Variational Automation where environment geometry is known (but object location is unknown), so the performance of CaP-X serves as an ablation of GaP with a single agent and without self-learning.  We evaluate VLA models $\pi_{0.5}$~\cite{black2025pi_} and MolmoAct2~\cite{fang2026molmoact2} with official checkpoints finetuned on the LIBERO training dataset. We also evaluate TipTop~\cite{shen2026tiptop}, a modular open-vocabulary planning system for robotic manipulation based on TAMP. We configure this from the github code with the default configuration of 128 particles for cuTAMP~\cite{shen2024differentiable} with a 60-second planning timeout.
We warm up the cuTAMP cache before measuring the execution time.  Rows 5 and 6 represent combinations of GaP with VLA models, where GaP generates an initial robot motion to center the wrist camera over the target object (bringing the VLA into distribution).

For GaP and CaP-X, we use Gemini-3.1-Flash-Lite for LLM agents and VLM with a temperature of 0.1.  
For Benchmarks I and II,  GaP does not perform self-learning because the first graph generated achieves high performance.

\subsection{Benchmarks I and II: Fulfill Grocery Orders and Pack Grocery Items}

\begin{table}[t]
  \centering
  \footnotesize
  
  \caption{\scriptsize \textbf{Success Rates for 5500 trials, 100 task instances per table cell. Results on Benchmarks I-a and II-a: Fulfill Grocery Orders and Pack Grocery Items}. The first two columns are the original LIBERO~\cite{liu2023libero} with negligible pose variation, and LIBERO-PRO with slightly larger pose variation~\cite{zhou2025liberopro} . We introduce the VA Grocery Order Fulfillment benchmark that includes larger object positional variance (X-Y 20$\times$20 {\scriptsize{$cm^2$}}), swapping target and basket positions (basket\_swap), permuting all item locations, and everything combined (mixed\_all). We also introduce the Grocery Packing benchmark that requires picking all objects and placing them in the basket. In rows 5 and 6, GaP navigates the robot wrist camera above the target object and then switches to MolmoAct2 and $\pi_{0.5}$ VLA policies.  }
  \label{tab:benchmark-banked}
  \resizebox{\linewidth}{!}{%
  \begin{tabular}{lcccccccc}
  \toprule
   & LIBERO & LIBERO-Pro & \multicolumn{4}{c}{Fulfill Grocery Orders} & \multicolumn{2}{c}{Pack Grocery Items} \\
  \cmidrule(lr){2-2}\cmidrule(lr){3-3}\cmidrule(lr){4-7}\cmidrule(lr){8-9}
  Method & object & object\_swap & X-Y 20$\times$20 {\small{$cm^2$}} & basket\_swap & permutation & mixed\_all & fixed & varied \\
  \midrule
  CaP-X   & - & 0.22 & 0.07 & 0.05 & 0.11 & 0.10 & 0.01 & 0.01 \\
  $\pi_{0.5}$               & 0.96 & 0.24 & 0.78 & 0.15 & 0.20 & 0.20 &   0.17   &   0.18   \\
  MolmoAct2                 & \textbf{0.97} & 0.43 & 0.90  & 0.26 & 0.10 & 0.20 &  0.18    &    0.18 \\
  TipTop   & 0.22      &   0.22   &  0.29    &   0.24   &   0.31   &   0.24   &   0.34   &     0.46    \\
  \midrule
  $\pi_{0.5}$ w/GaP         & 0.85 & 0.60 & 0.79 & 0.32 & 0.50 & 0.39 &  0.67    &  0.66    \\
  MolmoAct2 w/GaP           & 0.70 & 0.62 & 0.84 & 0.58 & 0.39 & 0.66 &  0.59    &    0.59  \\
  GaP                       & {0.95} & \textbf{0.95} &  \textbf{0.95} & \textbf{0.97} & \textbf{0.93} & \textbf{0.97} &   \textbf{0.99 }  &  \textbf{0.98}    \\
  \bottomrule
  \end{tabular}%
  }
  \vspace{-10pt}
\end{table}

\subsubsection{Simulation Results}
\textbf{GaP achieves significantly better positional robustness compared to baselines.} As shown in Table 1, while baseline VLA models ($\pi_{0.5}$, MolmoAct2) achieve high success rates on LIBERO-object without pose variation, their performance drops as low as 0.20 for LIBERO-PRO with positional variance. In contrast, GaP maintains robust success rates across all variations. We find that TipTop~\cite{shen2026tiptop} cannot generate motion plans for many of these benchmark task instances because M2T2~\cite{yuan2023m2t2}, cuRobo~\cite{sundaralingam2023curobo}, and cuTAMP~\cite{shen2024differentiable} cannot plan feasible solutions.

\begin{wraptable}{r}{0.32\columnwidth}
  \centering
  \vspace{-\baselineskip}  %
  \tiny
  \caption{\scriptsize \textbf{Physical Benchmark on Benchmarks I-b, II-b, and III-b}. We report success rate and completion time. Sub-tasks ($\rightarrow$) are reported per benchmark. On average, the completion time of single pick-and-place for TipTop is 95 seconds, and GaP is 67 seconds. }
  
  \label{tab:benchmark-transposed}
  \resizebox{\linewidth}{!}{%
  \begin{tabular}{lrr}
  \toprule
  Success Rate & TipTop & GaP \\
  \midrule
  Fulfill Grocery Orders            & 8/25   & \textbf{25/25} \\
  \quad$\rightarrow$ object        & 3/5   & 5/5 \\
  \quad$\rightarrow$ 20x20         & 2/5   & 5/5 \\
  \quad$\rightarrow$ basket swap   & 1/5   & 5/5 \\
  \quad$\rightarrow$ tall basket   & 0/5  & 5/5 \\
  \quad$\rightarrow$ permutation   & 2/5   & 5/5 \\
  \midrule
  Pack Grocery Items                & 10/30   & \textbf{28/30} \\
  \midrule
  Make Popcorn                &  0/20  & \textbf{18/20} \\
  \bottomrule
  \end{tabular}%
  \label{tab:real}
  }
  \vspace{-15pt}
  \vspace{-\baselineskip} 
\end{wraptable}
\textbf{VLA policies can benefit from GaP.} Performance for $\pi_{0.5}$ w/GaP and MolmoAct2 w/GaP suggest that GaP can enhance robustness of existing VLA models as shown in rows 5 and 6 of Table 1. As VLA policies struggle with positional perturbation of target objects, GaP uses the wrist camera and interactive perception skills to reliably identify the target object and uses linear cartesian motion skills to navigate the arm and wrist camera to a pre-grasp pose above the target object. The GaP computation graph then hands over execution to the VLA policy, staging it from a more favorable observation and producing more than a twofold improvement in success rates. %

\subsubsection{Real Physical Experiment Results}
\textbf{GaP transfers well from Sim to Real}. As shown in Table~\ref{tab:real}, 
for the {Fulfill Grocery Orders} benchmark, GaP achieves a perfect success rate of 25/25 (100\%), while TipTop succeeds in only 8/25 trials. 
Both GaP and TipTop accurately identify the objects. However, TipTop~\cite{shen2026tiptop} fails because M2T2~\cite{yuan2023m2t2}, cuRobo~\cite{sundaralingam2023curobo}, and cuTAMP~\cite{shen2024differentiable} combined cannot find feasible motion plans for cubic-shaped objects, tall baskets, or varied object orientations.
In terms of execution time, GaP perceives the item and basket in parallel using 14.1 seconds to generate oriented bounding boxes, and 36.4 seconds for descending, grasping, and transporting the items.

\subsection{Ablation Studies}
For Fulfilling Grocery Orders and Packing Grocery Items, the full GaP system achieves robust success rates (0.93–0.99). We ablate as follows: (1) Graphless generation forces the LLM to perfectly inline low-level service interfaces, such as method names and request schemas, from memory. Consequently, even when the high-level logic is correct, minor syntax or interface mismatch errors inevitably terminate the trial.  So replacing the structured graph output with a single LLM that generates raw Python scripts collapses the success rate to zero.
(2) Condensing GaP's specialized authoring agents into a single LLM also drops success rates to zero, with all trials failing static structural verification prior to execution, because a single LLM agent struggles to simultaneously manage global data flow, local logic, and verification, frequently introducing dangling references or node collisions. Furthermore, even with iterative feedback, a single LLM tends to oscillate between mistake classes without converging. 
(3) {Graph validation improves graph connectivity}. For each configuration in Fulfill Grocery Orders, some task segmentation agents emit a first-attempt subgraph that fails static type/edge checks and would crash at runtime, for example, some graphs wire the input and output of the {transport} phase in drop-pose computation to the release phase and other is in checkpoint verification. GaP's graph validation verifies the wiring and effectively ensures the graph edge connectivity.

\subsection{Benchmark III a, b: Make Popcorn in sim and real}

\begin{wrapfigure}{r}{0.5\columnwidth}
\centering
\vspace{-\baselineskip}
\includegraphics[width=\linewidth]{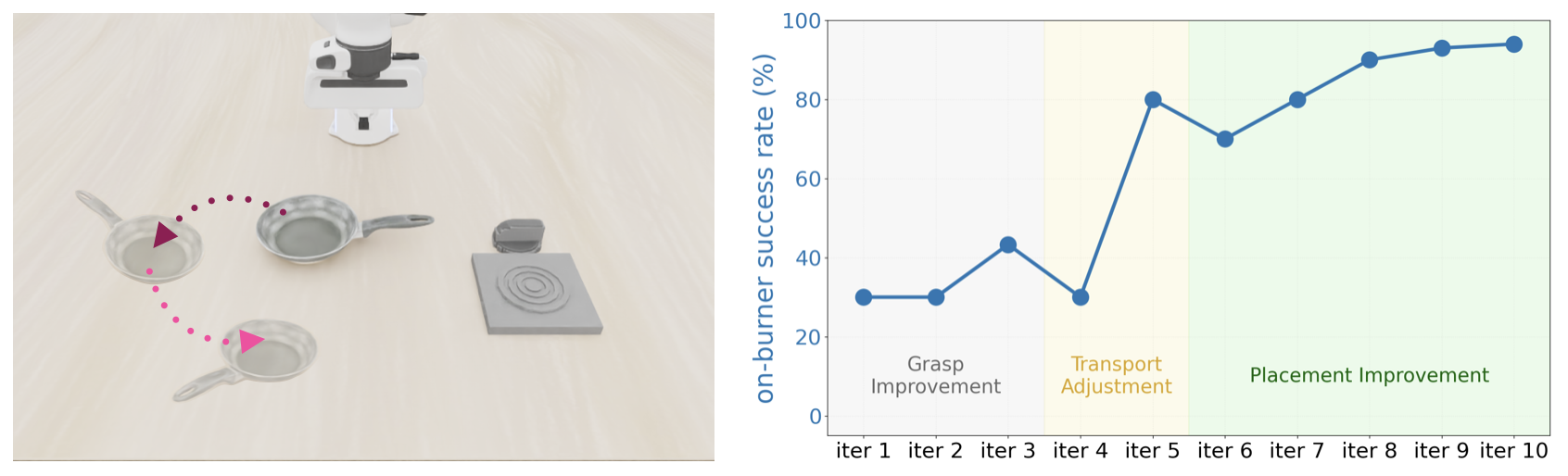}
\caption{\scriptsize \textbf{Self-Learning for Making  Popcorn  Benchmark.}
Pan pose variations (left) drive a 10-iteration sequence of graph updates; (blue, left axis) iteration phases shaded by class of edit.}
\label{fig:reh_res}
\vspace{-10pt}
\end{wrapfigure}
The Make Popcorn task requires the robot to grasps the stove knob and rotate it to turn on the burner, then the robot must find and pick up the handle of the JiffyPop popcorn pan, place it on the stove burner, wait, and then turn off the stove.

GaP generates a graph using GraspGen~\cite{murali2025graspgen}, knob turning skills and other object localization and grasping skills from Benchmark 1 and 2. 
The initial graph is able to turn on and off the knob but unable to pick and place the pan properly with only a 33\% success rate. 

\textbf{Self-Learning significantly increases success rate.} As shown in the Fig.~\ref{fig:reh_res}, self-learning proceeds through three stages: (1) from iteration 1 to 3, the majority of failures come from pan grasping failure, where the simulator feedback reports that the Franka gripper is not in contact with the object pan, so GaP replaces the GraspGen grasp skill with a skill that mixes GraspGen with an oriented bounding box grasp planner.
(2) In iteration 4, GaP  recognizes that the pan should be grasped by the handle, thereby adjusting the perception algorithm prompts to localize the pan handle.  
(3) In iterations 4-8, GaP finetunes the offset of pan placement due to the changed grasping strategy to align the pan with the stove surface. 

The resulting policy achieves 94\% in simulation with variations in pan position and orientation, and achieves a 90\% (18/20) success rate in real physical trials as shown in Table \ref{tab:real}. 
The two physical failures resulted from one inverse kinematics (IK) error during linear Cartesian motion, and one misgrasp caused by accumulated kinematic error.

\subsection{Self-Learning to Reduce Cycle Time}
\begin{figure}[t]
    \centering
    \includegraphics[width=0.88\linewidth]{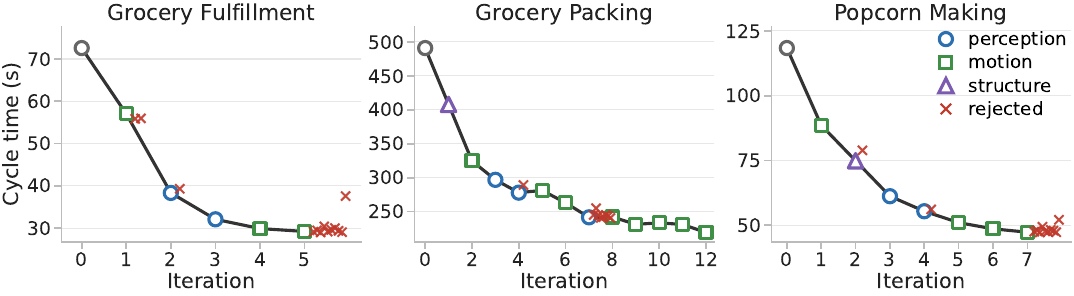}
    \caption{\scriptsize \textbf{Self-Learning for Cycle Time on Benchmarks I--III.} Per-iteration cycle time (left axis) and success rate (right axis). Accepted graph updates are colored by edit class: perception ({\color{NavyBlue}blue}), motion parameters ({\color{ForestGreen}green}), and graph topology ({\color{Plum}purple}); updates that degrade success rate or cycle time on 100 fresh task instances are {\color{purple}rejected}.}
    \label{fig:cycle_time}
\end{figure}
\begin{table}[t]
\centering
\scriptsize
\caption{\scriptsize \textbf{Success rate and throughput before and after self-learning on Benchmarks I--III (simulation).} TiPToP w/ MORSL replaces TiPToP's perception and grasp planning with MORSL skills.}
\label{tab:cycle-time}
\resizebox{0.6\linewidth}{!}{%
\begin{tabular}{lcccc|cc}
\toprule
& \multicolumn{4}{c|}{\textbf{Success Rate} $\uparrow$} & \multicolumn{2}{c}{\textbf{Throughput (cycles/hr)} $\uparrow$} \\
\cmidrule(lr){2-5} \cmidrule(l){6-7}
\textbf{Benchmark} &
\textbf{TiPToP} &
\textbf{\makecell{TiPToP \\ w/ MORSL}} &
\textbf{\makecell{GaP \\ (Initial)}} &
\textbf{\makecell{GaP \\ (Learned)}} &
\textbf{\makecell{GaP \\ (Initial)}} &
\textbf{\makecell{GaP \\ (Learned)}} \\
\midrule
I. Fulfill Grocery Orders & 0.24 & 0.53 & \textbf{0.97} & \textbf{0.97} & 49.7 & \textbf{124.1} \\
II. Pack Grocery Items    & 0.34 & 0.60 & \textbf{0.97} & \textbf{0.97} & 7.3 & \textbf{16.2} \\
III. Make Popcorn    & 0.00 & 0.00 & 0.33 & \textbf{0.94} & 30.5 & \textbf{76.6} \\
\bottomrule
\end{tabular}%
}
\vspace{-10pt}
\end{table}
Since GaP saturates the success rate on Benchmarks I--III, we additionally run self-learning with the throughput term of $\mathcal{J}$ to reduce cycle time. As shown in Fig.~\ref{fig:cycle_time} and Table~\ref{tab:cycle-time}, the LLM agent reduces cycle time by up to \textbf{57\%} with no success rate trade-off. It adjusts perception algorithms (e.g., using a cheaper detector for the basket), tunes motion parameters (e.g., control settling steps), and modifies graph topology (e.g., reusing perception results across nodes); proposals that degrade success rate or cycle time on 100 fresh task instances are rejected by the internal simulation.
Table~\ref{tab:cycle-time} also reports TiPToP equipped with MORSL perception and grasp planning skills: TAMP benefits from the higher-quality MORSL skills, while GaP further improves performance with flexible motion primitives and self-learning.
In terms of practical cost, graph generation takes fewer than 30 LLM calls to Gemini (about \$0.01 and 1~s latency); self-learning wall-clock time is dominated by simulation (about 15 minutes per iteration and up to 150 minutes in total for Make Popcorn); and no agent is involved once the graph is deployed.

\subsection{Benchmark IV: Insert USB-C Cables (Real)}

\begin{figure}[htbp]
    \centering
    
    \begin{minipage}{0.48\textwidth}
        \centering
        \includegraphics[width=0.9\linewidth]{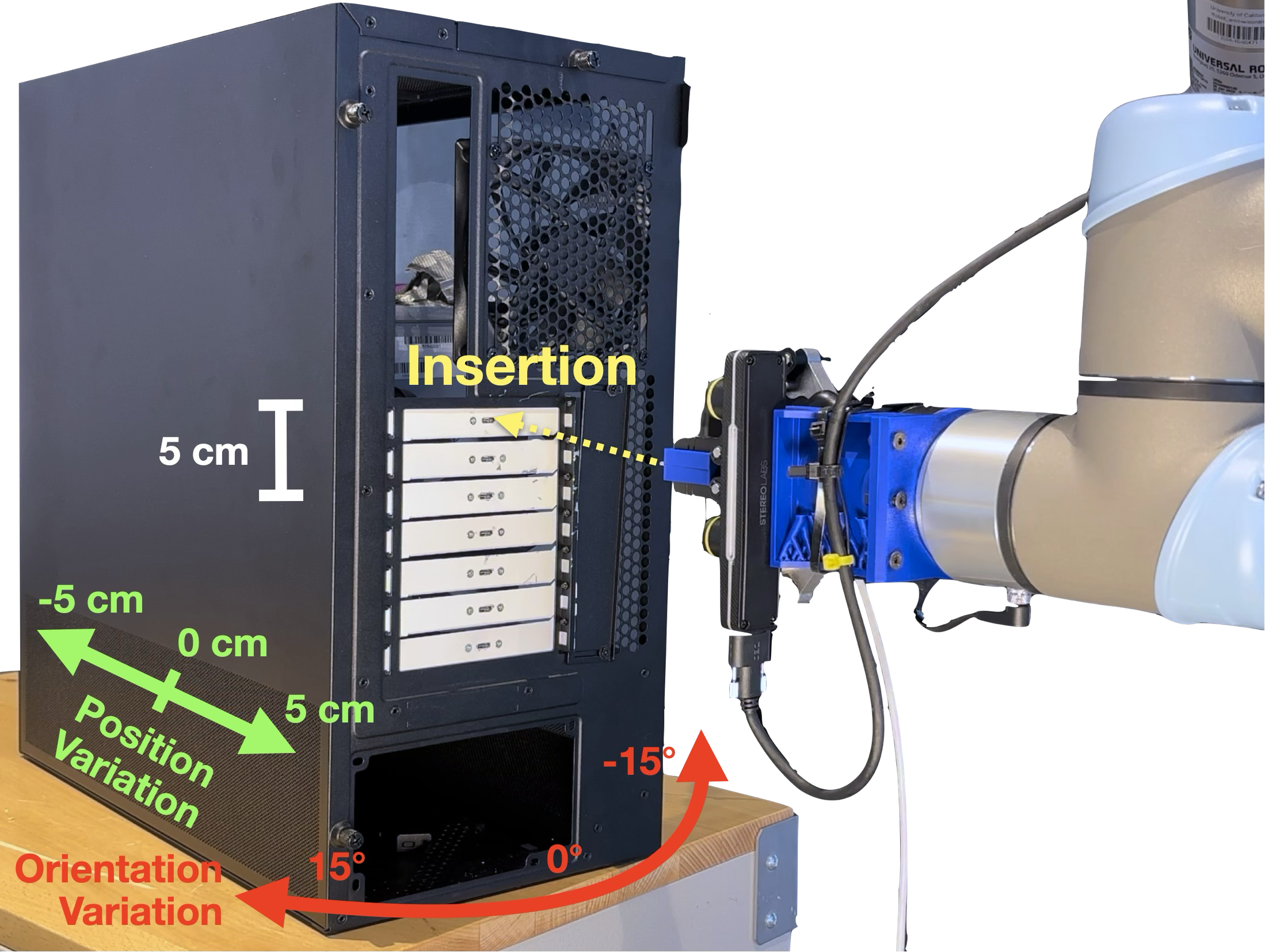}
        \captionof{figure}{Seven-ports cable insertion setup.}
        \label{fig:usb-insert-steup}
    \end{minipage}
    \hfill
    \begin{minipage}{0.48\textwidth}
        \centering
        \captionof{table}{\scriptsize \textbf{USB-C Cable insertion variations and results over 130 total insertion trials}. Success rates (SR) and execution time in seconds (ET). $^*$Excludes extraction phase.}
        \label{tab:benchmark-insertion}
        \scriptsize
\setlength{\tabcolsep}{4pt}
\renewcommand{\arraystretch}{0.95}
\begin{tabular}{lccc}
\toprule
\textbf{Variation} & \textbf{SR (Per trial)} & \textbf{SR (All)} & \textbf{ET (All)} \\
\midrule
\multicolumn{4}{c}{\textit{Insertion Goal Condition}} \\
\midrule
Individual port$^*$ & 5/5   & -   & 26.0 \\
Ascending order     & 32/35 & 2/5 & 247.0 \\
Descending order    & 32/35 & 4/5 & 223.2 \\
Odd index port      & 20/20 & 5/5 & 136.8 \\
Even index port     & 14/15 & 4/5 & 104.6 \\
\midrule
\multicolumn{4}{c}{\textit{Position/Orientation}} \\
\midrule
-$15^\circ$$^*$     & 4/5   & -   & 24.0 \\
$15^\circ$$^*$      & 5/5   & -   & 25.6 \\
-$5$ cm$^*$         & 5/5   & -   & 29.4 \\
$5$ cm$^*$          & 4/5   & -   & 28.0 \\
\bottomrule
\end{tabular}

    \end{minipage}
    
\end{figure}

For cable insertion benchmark evaluation, we utilize the computation graph generated by GaP, which operates via four modular ROS execution nodes: \textit{align to port}, \textit{touch port}, \textit{insert}, \textit{extract}. The robot first aligns to establish contact, generating a 1~$\times$1~$cm^2$ grid of insertion candidates. The \textit{insert} node then evaluates these positions, requiring $>$3~mm of progress at $<$10.0~N of force, and confirms successful insertion when the depth exceeds 6~mm at 30.0~N. Finally, the \textit{extract} node safely pulls the cable using a 2.0~Hz wiggling motion while maintaining lateral forces below 20.0~N. This entire policy requires approximately 30~seconds per insertion.

\begin{figure}
    \centering
    \includegraphics[width=1\linewidth]{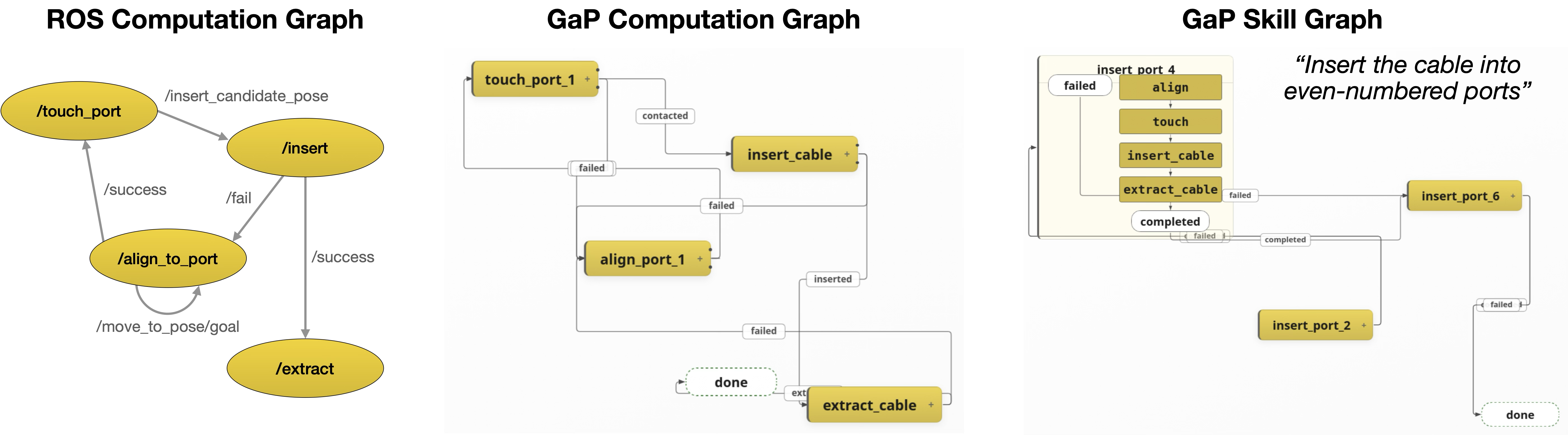}
    \caption{Left: ROS computation graph that is hand-engineered using traditional ROS nodes and topics. Middle: GaP can leverage nodes in ROS and generate a compatible graph using atomic skill sets, align to port, touch port, insert, and extract. Right: GaP can also use these nodes inside a subgraph for a long-horizon task, such as inserting the cable into even-numbered ports.}
    \label{fig:usb-insert-graph}
\end{figure}

\textbf{GaP provides flexibility to integrate with ROS nodes for robust, accurate task completion.} Across 130 insertion trials, GaP achieved a 0.93 success rate (121/130). A detailed breakdown is provided in Table~\ref{tab:benchmark-insertion}. GaP generated the correct workflows for all five text prompts, successfully handling long-horizon tasks with variations in position and orientation. Through this integration with ROS, GaP ensures both robust task execution and strong generalization across diverse positional variations and goal conditions. Some of the corresponding graphs generated by GaP are shown in Figure~\ref{fig:usb-insert-graph}.

\subsection{Benchmark V: Wash Crates (Simulation)}
\begin{wrapfigure}{r}{0.4\textwidth}
    \centering
    \vspace{-\baselineskip}
    \captionof{table}{\scriptsize \textbf{Wash Crates benchmark, 150 trials.} Success rate (SR), average cycle duration (s), and sustained throughput (successes/hr) over a 3-hour continuous run.}
    \label{tab:crate-wash}
    \scriptsize
    \resizebox{\linewidth}{!}{%
    \begin{tabular}{lccc}
    \toprule
    Method & SR & Avg(Dur) & Throughput\\
    \midrule
    Hand-Engineered & 0.99 & 176 & 19 \\
    GaP             & 0.95 & 179 & 18 \\
    \bottomrule
    \end{tabular}%
    }
    \vspace{6pt}
    \includegraphics[width=0.95\linewidth]{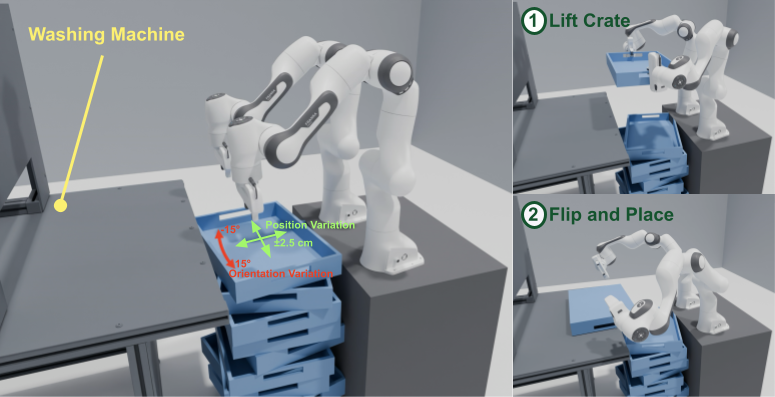}
    \caption{Industrial crate-washing setup.}
    \label{fig:crate-wash-setup}
    \vspace{-10pt}
\end{wrapfigure}
We compare GaP with a hand-engineered execution graph authored by an expert, representing the conventional manual engineering effort required in such settings. We run both policies over 150 trials sampled from the pose distribution, and additionally measure sustained throughput over a 3-hour continuous execution window.

\textbf{GaP closely matches an expert hand-engineered solution.} As reported in Table~\ref{tab:crate-wash}, GaP achieves a $0.953$ ($143/150$) success rate against the hand-engineered graph's $0.987$ ($148/150$), with similar average cycle times ($179.13$~s vs.\ $176.47$~s). Over the 3-hour throughput run both policies attempted $59$ trials, of which GaP completed $55$ and the hand-engineered graph $58$, corresponding to $18.33$ and $19.33$ successes per hour, respectively. These results suggest that GaP can autonomously produce coordinated bimanual policies whose reliability and throughput approach those of a manually hand-tuned baseline.
\section{Conclusion}

Agentic robotics has potential to build on rapidly-advancing LLM models to provide a bridge between Good Old Fashioned Engineering (GOFE) and model-free VLA policies.  As shown in the Ablation experiments, GaP's graph-structured agentic robotics facilitates modularity, multi-agent integration, and self-learning.  As robot applications become more complex and skill libraries grow, more sophisticated agent harnessing will be required.

\textbf{Limitations.} Although in experiments, GaP improves success rates significantly over baselines, execution reliability is not yet at industry levels and additional self-learning and parameter tuning is required.  Similarly, GaP execution times are still well below industry standards of 500 units per hour (7 seconds per instance);  more work is required to reduce VLM inference requests and IK motion planning time during execution.  Also, the 8 VA benchmarks focus on quasi-static pick-and-place operations -- only Cable Insertion (IV) requires force sensing.  More work is required to apply GaP to tasks with deformable objects, dynamic forces, and moving targets.  

\bibliography{example}  %

\appendix
\clearpage

\appendix

\section*{Appendix}

\addcontentsline{toc}{section}{Appendix}

\startcontents[appendix]

\printcontents[appendix]{}{1}{\section*{Contents}}

\clearpage

\counterwithin{figure}{section} %
\setcounter{figure}{0}          %

\section{More Related Work}

\textbf{Definition for Generalist Robotics (GR)}
In FA, variations in the environment and object shape and pose are minimal. For GR, on the other hand, a robot performs a variety of different tasks (e.g., placing groceries into a refrigerator, or 
domestic cleaning, folding, kitchen pick-and-place) in different homes, where environments vary considerably and objects have variable geometry and highly variable initial poses. Recent robot learning research predominantly targets highly unstructured Generalist Robotics (GR), such as household robots executing diverse chores across novel environments using a single generalist model-free VLA policies~\cite{levine2016end, chi2025diffusion, kim24openvla, pi0_2024, black2025pi_}.

\textbf{Self-Improving Agentic Workflows}.
Outside of robotics, Large Language Models (LLMs) have demonstrated exceptional capabilities in dynamic coding, complex software integration, and API orchestration \cite{wu2024introducing, team2026qwen3, hou2025model, openai2024gpt4o, anthropic2024claude35, google2025gemini25}. By embedding these capabilities into structured agentic workflows, LLMs can autonomously synthesize solutions for open-ended tasks~\cite{zhang2025aflow}. A foundational example is Voyager \cite{wang2023voyager}, which uses LLMs to iteratively write, refine, and execute Minecraft game playing skills to continually expand a curated library of reusable skills. In these frameworks, a ``skill'' is typically defined as a discrete function with strict semantic contracts. Recent literature has begun to explore the iterative optimization of agentic systems through language-based failure reasoning \cite{zehle2026promptolution} and multi-agent prompt optimization frameworks \cite{agrawal2025gepa, lee2025feedback}.
To improve code generation quality and performance, CaP-X~\cite{fu2026cap} used a VLM to provide feedback before and after execution (i.e. Visual Differencing), but the VLM can suffer from hallucinations and cannot handle geometric and numerical information such as motion feasibility. 
Building on the structure of Blox-Net~\cite{goldberg2025bloxnet} and BrickGPT~\cite{pun2025brickgpt}, which uses physical experiments to improve LLM-generated task plans, GaP uses multiple LLM agents to generate a robot computation graph and iteratively improve it using simulation experiments.

\section{Integrating ROS with GaP in Cable Insertion Benchmark}

We demonstrate how GaP can integrate the ROS through a cable insertion benchmark. Four nodes (\textit{align to port}, \textit{touch port}, \textit{insert}, \textit{extract}) are exposed to GaP through an intermediate skill interface that manages and triggers the ROS nodes sequentially. Upon receiving a call from the Behavior Graph, the skill interface instantiates a temporary ``orchestrator'' node to handle communication with the target ROS node, waiting until it receives a success, failure, or data signal. This execution status is then returned to the Behavior Graph, which determines the next workflow step and triggers subsequent ROS nodes accordingly.

\begin{figure}[b!]
    \centering
    \includegraphics[width=1\linewidth]{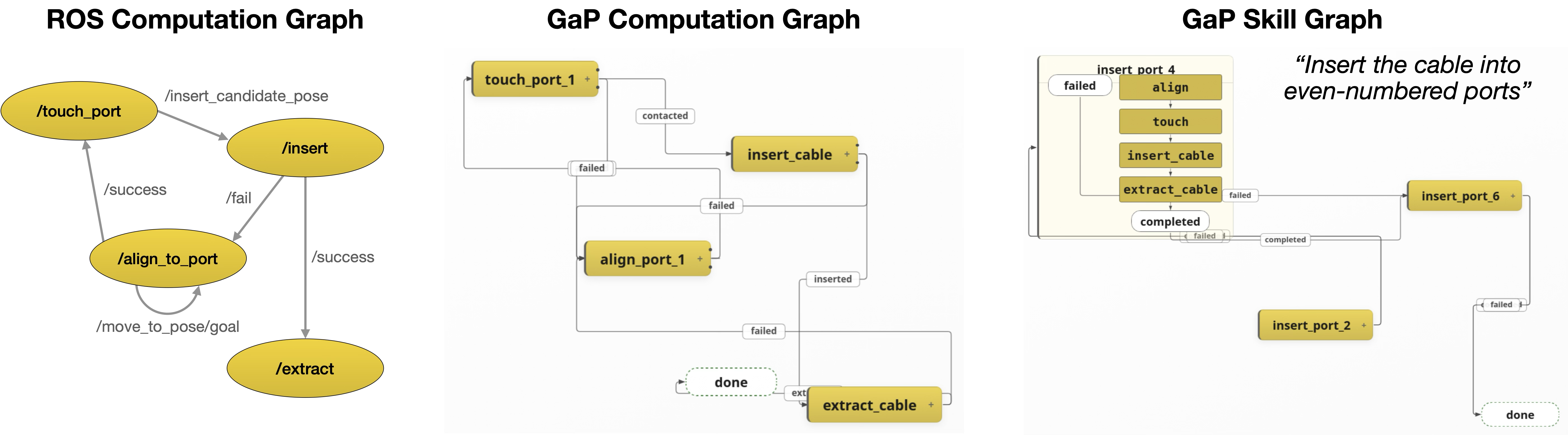}
    \caption{Left: ROS computation graph that is hand-engineered using traditional ROS nodes and topics. Middle: GaP can leverage nodes in ROS and generate a compatible graph using atomic skill sets, align to port, touch port, insert, and extract. Right: GaP can also use these nodes inside a subgraph for a long-horizon task, such as inserting the cable into even-numbered ports.}
    \label{fig:usb-insert-graph}
\end{figure}

\textbf{GaP provides flexibility to integrate with ROS nodes for robust, accurate task completion.} In our experiment, we create a computation graph using GaP with five different prompts: 1. individual port insertion ``Insert the cable into port X.'' (X: 1,2,4,5,7), 2. ``Insert the cable into ports in ascending order'', 3. ``Insert the cable into ports in descending order'', 4. ``Insert the cable into odd-indexed ports'', 5. ``Insert the cable into even-indexed ports''. Across 130 insertion trials, GaP achieved a 0.93 success rate (121/130). A detailed breakdown is provided in Table~\ref{tab:benchmark-insertion}. Some of the corresponding graphs generated by GaP are shown in Figure~\ref{fig:usb-insert-graph}. The framework successfully connected atomic skills to complete the core insertion task and demonstrated the capacity to generate subgraphs using these skills to handle extended long-horizon tasks. The GaP-created graph is comparable to an expert-engineered baseline graph built using traditional ROS nodes and topics (see the side-by-side comparison in Figure~\ref{fig:usb-insert-graph} left and middle). 

We identified three main failure modes in our insertion policy. First, when the initial vision-based insertion candidate fails, the system falls back to a contact-based search within a 1~$\times$~1~$cm^2$ area in 2~mm steps. If the initial vision estimate falls outside this search zone, the brute-force probing ultimately fails. Second, successful insertion requires precise alignment with the housing; inaccurate depth estimation occasionally results in only a partial insertion. Lastly, variable lighting conditions or shadows cast by the robot and housing can cause the initial port detection to fail.

\section{Sample Generated Graphs}

\subsection{Fulfill Grocery Orders}

\begin{figure}[H]
    \centering
    \includegraphics[width=0.98\linewidth]{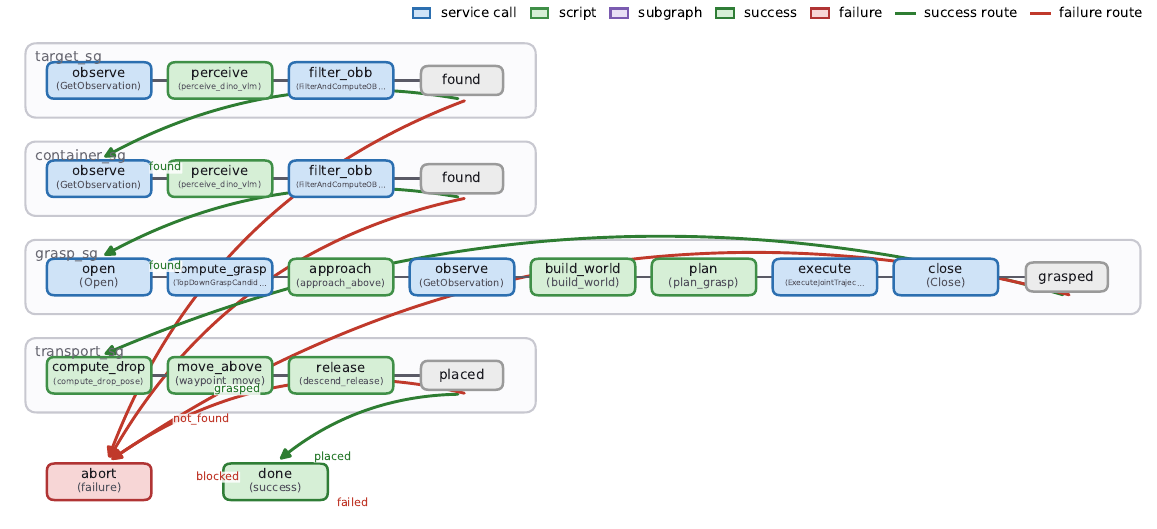}
    \label{fig:placeholder}
\end{figure}

\subsection{Pack Grocery Items}

\begin{figure}[H]
    \centering
    \includegraphics[width=0.98\linewidth]{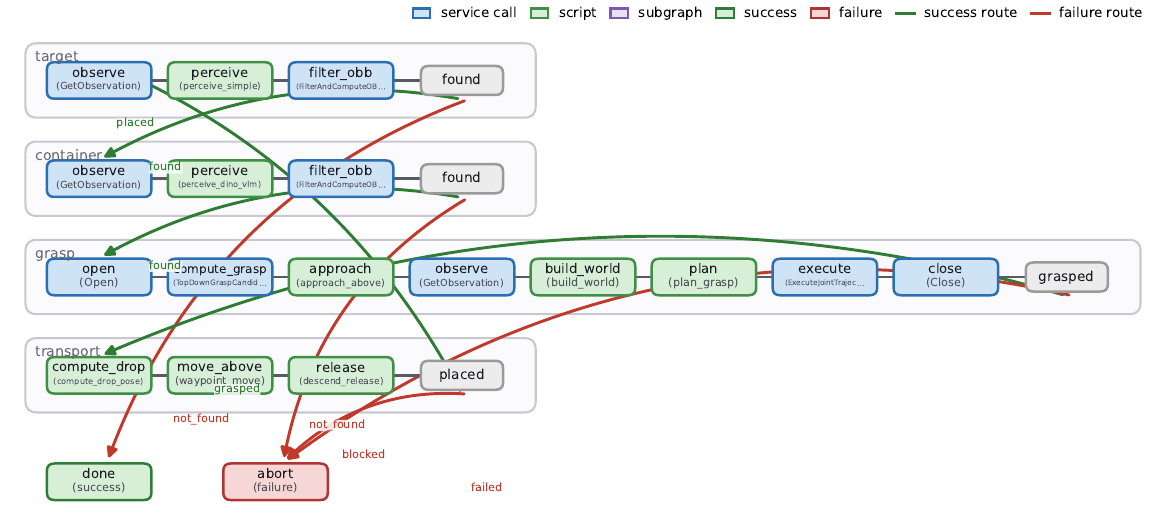}
    \label{fig:placeholder}
\end{figure}

\subsection{Fulfill Grocery Orders with VLA Policy}

\begin{figure}[H]
    \centering
    \includegraphics[width=0.9\linewidth]{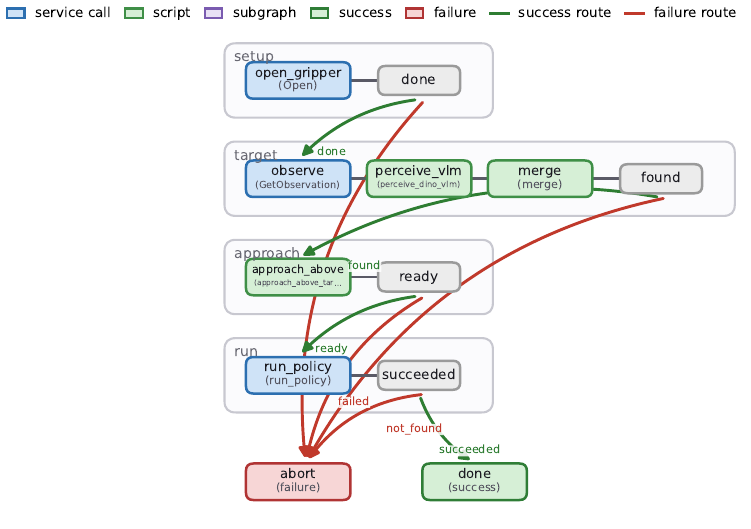}
    \label{fig:placeholder}
\end{figure}

\subsection{Wash Crates}

\begin{figure}[H]
    \centering
    \includegraphics[width=\linewidth]{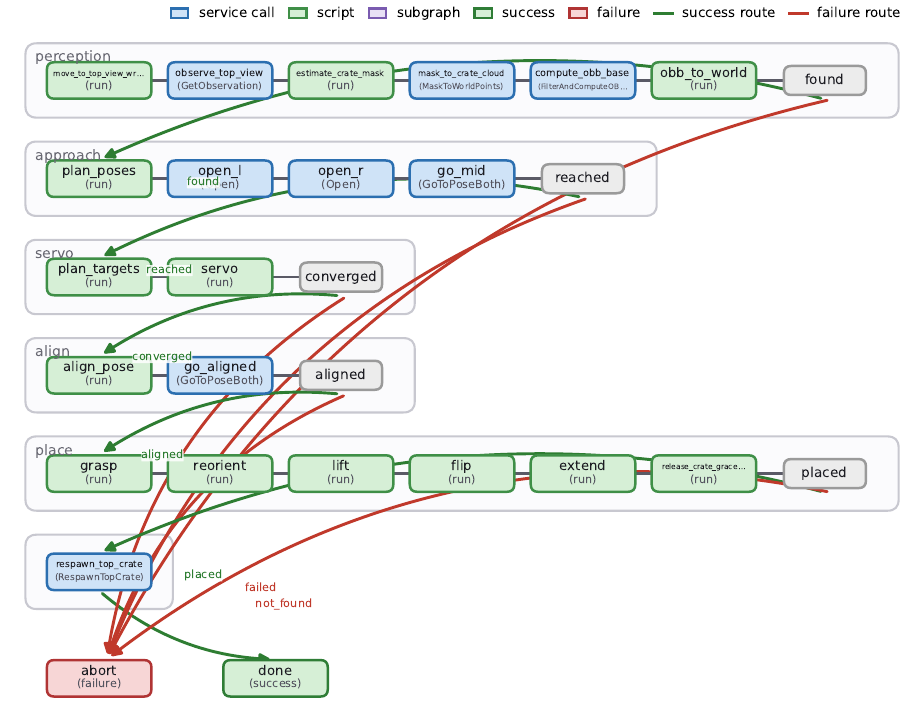}
    \label{fig:placeholder}
\end{figure}

\subsection{Make Popcorn}

\begin{figure}[H]
    \centering
    \includegraphics[width=\linewidth]{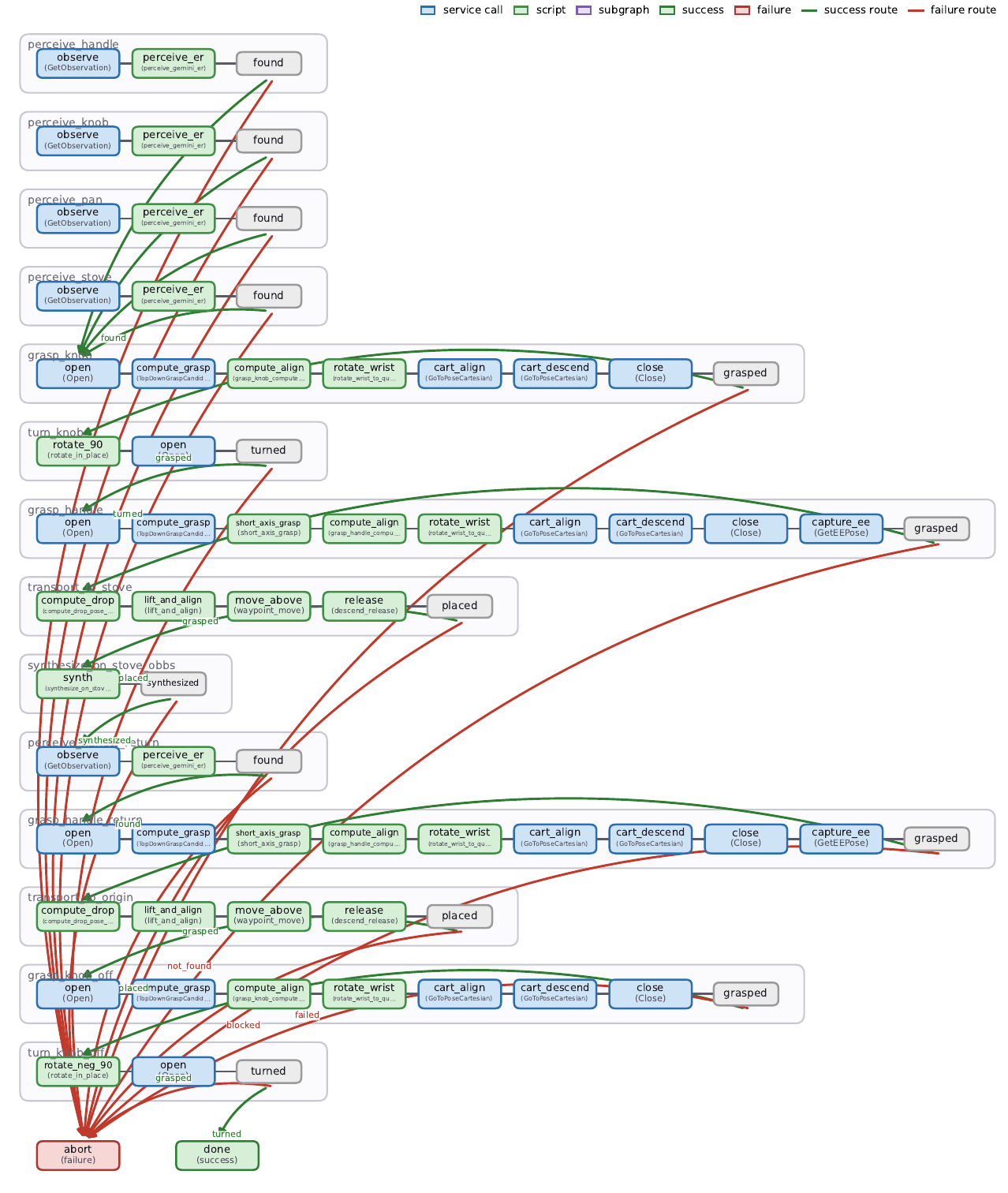}
    \label{fig:placeholder}
\end{figure}

\section{MORSL Library}

The system is built from a single kind of component, the \emph{skill}.
Every skill is declared by the same metadata schema --- a name, a
description, and its typed inputs and outputs. Higher-level
\emph{composite} skills are materializable subgraphs authored by
composing other skills; \emph{atomic} skills are single classes; and
\emph{primitive} skills are individual gRPC methods, each wrapping one
model invocation or geometric routine. All skills share a common type
system (\texttt{Se3Pose}, \texttt{OrientedBoundingBox},
\texttt{PointCloud}, \texttt{Mask}, \texttt{Image}, \texttt{Trajectory},
\texttt{JointState}, $\dots$), which is what lets them compose freely. In
the input lists below, defaults are given in parentheses and the
universal caller-context field is omitted.

\subsection{Composite and atomic skills}

\paragraph{Perception.}

\skillcard{perception\_single}
{\texttt{object\_name} (literal)}
{\texttt{<name>\_obb}: OrientedBoundingBox, \texttt{<name>\_mask}: Mask, \texttt{<name>\_cloud}: PointCloud}
{Fast single-path detection: broad detect + a VLM crop tournament +
segmentation + depth-to-3D fusion, closed by an OBB fit. Best for
uncluttered scenes with visually distinct targets.}

\skillcard{perception\_multi}
{\texttt{object\_name} (literal)}
{\texttt{<name>\_obb}: OrientedBoundingBox, \texttt{<name>\_mask}: Mask}
{Three detectors run on the same observation and a VLM disambiguator picks
the best mask on disagreement. Most robust, but slower; single-shot only.}

\skillcard{perception\_any}
{\texttt{object\_name} (literal)}
{\texttt{<name>\_obb}: OrientedBoundingBox, \texttt{<name>\_mask}: Mask, \texttt{<name>\_cloud}: PointCloud}
{Lightweight one-shot set-of-marks pick. Returns ``not found'' cleanly
when the VLM replies ``none'', the right choice for clean-all-items loops.}

\skillcard{perception\_subpart}
{\texttt{parent\_prompt}, \texttt{subpart\_prompt} (literals)}
{\texttt{<name>\_obb}: OrientedBoundingBox, \texttt{<name>\_mask}: Mask, \texttt{<name>\_cloud}: PointCloud}
{Hierarchical perception: detect the parent, crop to its box, segment the
named subpart inside the crop, uncrop and fuse depth. For affordances that
occupy few full-image pixels (pan handle, drawer pull, mug rim).}

\skillcard{track\_object}
{text prompt; observation stream}
{a stream of tracked-mask snapshots}
{Long-running tracker: seeds on the first frame via text prompt, then polls
the observation stream and advances the tracker until the workflow signals
termination, freeing tracker state on exit.}

\paragraph{Grasping.}

\skillcard{grasp\_curobo\_obb}
{\texttt{target\_obb}: OrientedBoundingBox, \texttt{target\_mask}: Mask}
{\texttt{ee\_pose\_at\_grasp}: Se3Pose, \texttt{grasp\_pose}: Se3Pose}
{Collision-aware grasping: plans a collision-free trajectory to one of
several top-down OBB grasp candidates. The default when the OBB centroid
is graspable.}

\skillcard{grasp\_moe}
{\texttt{target\_obb}: OrientedBoundingBox, \texttt{target\_mask}: Mask, \texttt{target\_cloud}: PointCloud}
{\texttt{ee\_pose\_at\_grasp}: Se3Pose}
{Mixture-of-experts grasping: a diffusion sampler and a dense OBB sweep
feed one discriminator; the top-K union is planned. Use when the OBB
centroid is not the graspable point (bowl rim, handle, off-center).}

\skillcard{grasp\_short\_axis}
{\texttt{target\_obb}: OrientedBoundingBox, \texttt{target\_mask}: Mask, \texttt{base\_obb}: OrientedBoundingBox (optional)}
{\texttt{ee\_pose\_at\_grasp}: Se3Pose, \texttt{grasp\_pose}: Se3Pose}
{Deterministic geometry-locked grasp: the finger axis snaps to the OBB's
shorter horizontal axis so the jaws close across the narrow dimension of
an elongated target. An optional base OBB slides the grasp along a handle.}

\skillcard{grasp\_direct\_ik}
{\texttt{target\_obb}: OrientedBoundingBox}
{(grasp completion only)}
{Planner-free grasping: pre-rotate to the grasp orientation above the
target, descend straight down, close. For uncluttered scenes or platforms
without a motion planner.}

\paragraph{Transport, policy, and bimanual.}

\skillcard{transport\_to\_drop}
{\texttt{container\_obb}, \texttt{container\_mask}, \texttt{target\_obb}, \texttt{target\_mask}, \texttt{ee\_pose\_at\_grasp}}
{(placement completion only)}
{Moves a held object above a destination container and releases it. An
optional VLM-grounded step localizes a natural-language sub-region (``the
left compartment'') before the drop pose is computed.}

\skillcard{run\_policy}
{observation stream; \texttt{max\_windows}}
{(policy rollout completion)}
{Runs a learned VLA policy in closed loop, reading the observation stream
each window, until a VLM termination check fires or \texttt{max\_windows}
is reached.}

\skillcard{bimanual\_crate\_lift}
{(none --- waypoints are baked parameters)}
{\texttt{done}: bool, \texttt{final\_left\_world\_pos}, \texttt{final\_right\_world\_pos}}
{Two-arm crate lift: both arms side-grasp the crate's handle bars and raise
it in lock-step, in a fixed phase order
(open\,$\to$\,approach\,$\to$\,insert\,$\to$\,close\,$\to$\,lift\,$\to$\,settle).}

\subsection{Primitive skills (one per gRPC method)}

\paragraph{Detection and segmentation.}

\skillcard{grounding\_dino.Detect}
{\texttt{image}: Image; \texttt{text\_prompt} (period-separated phrases); \texttt{box\_threshold} (0.20); \texttt{text\_threshold} (0.20)}
{\texttt{detections}: [\,box: BoundingBox2D, label, score\,]}
{Zero-shot object detection from a text prompt; returns boxes, labels, and
confidence scores.}

\skillcard{owlvit.Detect}
{\texttt{image}: Image; \texttt{text\_queries}: string[]; \texttt{score\_threshold} (0.05)}
{\texttt{detections}: [\,box: BoundingBox2D, label, score\,]}
{Open-vocabulary object detection from a list of text queries; an
alternative to grounding\_dino.}

\skillcard{sam3.SegmentText}
{\texttt{image}: Image; \texttt{text\_prompt}}
{\texttt{masks}: Mask[], \texttt{scores}: float[], \texttt{boxes}: BoundingBox2D[]}
{Segment all instances matching a text description.}

\skillcard{sam3.SegmentPoint}
{\texttt{image}: Image; \texttt{pixel\_x}, \texttt{pixel\_y}}
{\texttt{masks}: Mask[], \texttt{scores}: float[]}
{Segment the object under a single pixel coordinate.}

\skillcard{sam3.SegmentBox}
{\texttt{image}: Image; \texttt{box}: BoundingBox2D; \texttt{pixel\_x}/\texttt{pixel\_y} + \texttt{use\_point} (false, optional foreground point)}
{\texttt{masks}: Mask[], \texttt{scores}: float[]}
{Segment within a bounding box, optionally refined by a foreground point.}

\skillcard{sam2.SegmentBox}
{\texttt{image}: Image; \texttt{box}: BoundingBox2D; \texttt{max\_masks} (0 = unlimited)}
{\texttt{masks}: Mask[], \texttt{scores}: float[]}
{Box-prompted segmentation (lighter SAM2 backend).}

\skillcard{sam2.SegmentPoint}
{\texttt{image}: Image; \texttt{pixel\_x}, \texttt{pixel\_y}}
{\texttt{masks}: Mask[], \texttt{scores}: float[]}
{Point-prompted segmentation (lighter SAM2 backend).}

\skillcard{sam3\_tracker.InitTracker}
{\texttt{image}: Image (first frame); one of \texttt{text} / \texttt{box} / point (\texttt{pixel\_x},\texttt{pixel\_y},\texttt{use\_point}); \texttt{object\_name}}
{\texttt{tracker\_id}, \texttt{initial\_mask}: Mask, \texttt{initial\_box}, \texttt{score}, \texttt{object\_present}}
{Initialize a stateful tracker session seeded with one frame and a prompt.}

\skillcard{sam3\_tracker.UpdateTracker}
{\texttt{tracker\_id}; \texttt{image}: Image (new frame)}
{\texttt{mask}: Mask, \texttt{box}, \texttt{confidence}, \texttt{object\_present}}
{Advance an existing tracker session by one frame.}

\skillcard{sam3\_tracker.CloseTracker}
{\texttt{tracker\_id}}
{(empty)}
{Free a tracker session; idempotent.}

\skillcard{sam3d.GenerateMesh}
{\texttt{object\_name}; \texttt{views}: [rgb, mask, intrinsics, camera\_pose]; \texttt{obb}: OrientedBoundingBox; \texttt{view\_selection}; \texttt{seed} (42); \texttt{output\_dir}}
{\texttt{mesh\_path}, \texttt{texture\_path}, \texttt{world\_pose}: Se3Pose, \texttt{scale}: Vec3, \texttt{chosen\_view\_index}, \texttt{num\_vertices}, \texttt{num\_faces}}
{Reconstruct a triangle mesh from an RGB image + mask, scaled and posed
into world frame via the OBB; writes an \texttt{.obj} for the rehearsal sandbox.}

\paragraph{Vision--language.}

\skillcard{molmo.PointPrompt}
{\texttt{image}: Image; \texttt{object\_name}}
{\texttt{pixel\_x}, \texttt{pixel\_y}, \texttt{found}}
{Point to a named object; returns pixel coordinates.}

\skillcard{molmo.Query}
{\texttt{prompt}; \texttt{image}: Image (optional); \texttt{use\_multiview}}
{\texttt{text}}
{Free-form visual question answering (Molmo backend).}

\skillcard{molmo.QueryYesNo}
{\texttt{prompt}; \texttt{image}: Image (optional); \texttt{use\_multiview}}
{\texttt{answer}: bool}
{Yes/no visual question answering (Molmo backend).}

\skillcard{vlm.Query}
{\texttt{prompt}; \texttt{image}: Image (optional)}
{\texttt{text}}
{Free-form visual question answering (configurable VLM backend).}

\skillcard{vlm.QueryYesNo}
{\texttt{prompt}; \texttt{image}: Image (optional)}
{\texttt{answer}: bool}
{Yes/no visual question answering (configurable VLM backend).}

\paragraph{Grasp generation.}

\skillcard{graspgen.PlanFromPointCloud}
{\texttt{target\_cloud}: PointCloud; \texttt{scene\_cloud} (optional); \texttt{num\_grasps} (200); \texttt{topk\_num\_grasps} (50); \texttt{grasp\_threshold} ($-1$); \texttt{remove\_outliers} (true)}
{\texttt{grasp\_poses}: Se3Pose[], \texttt{scores}: float[], \texttt{contact\_points}: Vec3[]}
{Diffusion-based 6-DoF grasp sampling from an object-segmented cloud;
ranked, world-frame, fingertip convention.}

\skillcard{graspgen.ScoreGrasps}
{\texttt{target\_cloud}: PointCloud; \texttt{candidate\_poses}: Se3Pose[]}
{\texttt{scores}: float[] (aligned to candidates)}
{Score caller-provided grasps with GraspGen's discriminator, so
OBB-sampled candidates are comparable with diffusion samples.}

\skillcard{graspnet.PlanFromDepth}
{\texttt{depth}: DepthImage; \texttt{intrinsics}: Matrix3x3; \texttt{segmentation}: Mask; \texttt{instance\_id}; knobs (\texttt{z\_min} 0.2, \texttt{z\_max} 2.0, \texttt{max\_retries} 7, $\dots$); optional wrist camera}
{\texttt{grasp\_poses}: Se3Pose[], \texttt{scores}: float[], \texttt{contact\_points}: Vec3[]}
{Contact-GraspNet 6-DoF grasp detection from a depth image + instance mask.}

\skillcard{graspnet.PlanFromPointCloud}
{\texttt{full\_cloud}: PointCloud; \texttt{segment\_cloud}: PointCloud; \texttt{instance\_id}; same knobs}
{\texttt{grasp\_poses}: Se3Pose[], \texttt{scores}: float[], \texttt{contact\_points}: Vec3[]}
{Contact-GraspNet grasp detection from precomputed clouds.}

\skillcard{m2t2.PlanFromPointCloud}
{\texttt{scene\_cloud}: PointCloud; \texttt{target\_cloud} (optional); \texttt{num\_points} (16384); \texttt{grasp\_threshold} (0.035); \texttt{max\_grasps} (200)}
{\texttt{grasp\_poses}: Se3Pose[], \texttt{scores}: float[], \texttt{contact\_points}: Vec3[]}
{M2T2 6-DoF grasp generation from a colored scene cloud; an alternative
candidate source.}

\paragraph{Motion planning and IK.}

\skillcard{curobo.PlanToGraspPoses}
{\texttt{world\_config}: WorldConfig; \texttt{start\_joint\_position}: JointState; \texttt{grasp\_poses}: Se3Pose[]; thresholds + collision knobs}
{\texttt{success}, \texttt{trajectory}: Trajectory, \texttt{goalset\_index}}
{Plan a collision-free trajectory to one of several grasp poses (goalset).}

\skillcard{curobo.PlanWithGraspedObject}
{\texttt{world\_config}; \texttt{start\_joint\_position}; \texttt{target\_pose}: Se3Pose; \texttt{object\_name}}
{\texttt{success}, \texttt{trajectory}: Trajectory}
{Plan a collision-free trajectory while holding a named grasped object.}

\skillcard{curobo.PlanLinear}
{\texttt{start\_pose}, \texttt{end\_pose}: Se3Pose; \texttt{start\_joint\_position}; \texttt{world\_config} (optional); \texttt{num\_waypoints} (20); \texttt{hold\_vec\_weight}}
{\texttt{success}, \texttt{trajectory}: Trajectory, \texttt{failure\_reason}}
{GPU-accelerated straight-line Cartesian motion via per-waypoint IK,
optionally constrained and collision-checked.}

\skillcard{curobo.PlanDirectedLinear}
{\texttt{start\_joint\_position}; \texttt{start\_pose}, \texttt{target\_pose}; \texttt{allowed\_axes}; \texttt{endpoint\_mode}; \texttt{orientation\_mode}}
{\texttt{success}, \texttt{trajectory}: Trajectory, \texttt{failure\_reason}}
{Constrained linear motion that holds selected Cartesian axes
(project-to-target / fixed-distance / orient-in-place).}

\skillcard{curobo.PlanGraspMotion}
{\texttt{start\_joint\_position}; \texttt{grasp\_pose}: Se3Pose; approach axis+distance (0.12); lift axis+distance (0.20)}
{\texttt{success}; \texttt{approach\_trajectory}, \texttt{grasp\_trajectory}, \texttt{lift\_trajectory}: Trajectory; \texttt{failure\_reason}}
{Three-phase grasp plan (free-space approach, constrained grasp,
constrained lift) so the caller can interleave gripper commands. (CuRobo v0.8.)}

\skillcard{curobo.SolveIK}
{\texttt{target\_pose}: Se3Pose; \texttt{seed\_config} (optional); \texttt{tcp\_offset}; \texttt{num\_seeds} (32)}
{\texttt{success}, \texttt{joint\_config}: JointState}
{Pure geometric IK for a single TCP pose (no world collision).}

\skillcard{curobo.PlanToPose}
{\texttt{target\_pose}: Se3Pose; \texttt{start\_joint\_position}; \texttt{world\_config} (optional); \texttt{tcp\_offset}}
{\texttt{success}, \texttt{trajectory}: Trajectory}
{Collision-aware trajectory to a single EE target pose.}

\skillcard{curobo.BatchGraspFeasibility}
{\texttt{world\_config}; \texttt{start\_state}: JointState; \texttt{grasp\_poses}: Se3Pose[]; \texttt{approach\_offset\_m} (0.10); \texttt{ignore\_obstacle\_names}}
{\texttt{feasible}: bool[], \texttt{grasp\_ik\_ok}: bool[], \texttt{approach\_ik\_ok}: bool[], \texttt{corridor\_collision\_fraction}: float[] (all aligned to input)}
{Per-pose scene-collision feasibility for a batch of grasp candidates;
filters blocked grasps before planning.}

\skillcard{curobo.ValidateJointTrajectoryRobot}
{\texttt{world\_config}; \texttt{trajectory}: Trajectory; collision knobs}
{\texttt{success}, \texttt{failure\_reason}, \texttt{first\_collision\_waypoint}, \texttt{collision\_status\_detail}}
{Validate joint waypoints against world + self-collision.}

\skillcard{curobo.ValidateJointTrajectoryGrasped}
{\texttt{world\_config}; \texttt{trajectory}; \texttt{object\_name}; collision knobs}
{\texttt{success}, \texttt{failure\_reason}, \texttt{first\_collision\_waypoint}, \texttt{collision\_status\_detail}}
{As above, with a grasped object attached at the first waypoint.}

\skillcard{pyroki.SolveIK}
{\texttt{target\_pose}: Se3Pose; \texttt{target\_link\_name} (panda\_hand); \texttt{prev\_config} (optional); \texttt{tcp\_offset} ($0,0,-0.1$)}
{\texttt{joint\_config}: JointState, \texttt{success}}
{CPU differentiable IK for a target EE pose; the planner-free IK fallback.}

\skillcard{pyroki.PlanLinear}
{\texttt{start\_pose}, \texttt{end\_pose}: Se3Pose; \texttt{num\_waypoints} (20); \texttt{dt} (0.02)}
{\texttt{trajectory}: Trajectory, \texttt{success}}
{CPU linear Cartesian trajectory between two poses.}

\paragraph{Geometry.}

\skillcard{geometry\_svc.FilterNoise}
{\texttt{point\_cloud}: PointCloud; \texttt{eps} (0.005); \texttt{min\_samples} (10)}
{PointCloud}
{DBSCAN noise filtering of a point cloud.}

\skillcard{geometry\_svc.ComputeOBB}
{\texttt{point\_cloud}: PointCloud}
{OrientedBoundingBox}
{Fit an oriented bounding box to 3D points.}

\skillcard{geometry\_svc.FilterAndComputeOBB}
{\texttt{point\_cloud}: PointCloud; \texttt{eps} (0.005); \texttt{min\_samples} (10)}
{OrientedBoundingBox}
{FilterNoise + ComputeOBB in one round trip.}

\skillcard{geometry\_svc.TopDownGraspFromOBB}
{\texttt{obb}: OrientedBoundingBox; \texttt{z\_offset}}
{Se3Pose}
{A single top-down grasp pose from an OBB.}

\skillcard{geometry\_svc.TopDownGraspCandidates}
{\texttt{obb}: OrientedBoundingBox; \texttt{z\_offset}}
{\texttt{poses}: Se3Pose[] (primary + 90$^\circ$-rotated)}
{Candidate top-down grasps so a planner can pick a reachable one.}

\skillcard{geometry\_svc.SelectTopDownGrasp}
{\texttt{grasp\_poses}: Se3Pose[]; \texttt{gripper\_position}: Vec3}
{Se3Pose}
{Pick the most top-down (and nearest) grasp from candidates.}

\skillcard{geometry\_svc.FrontGraspFromOBB}
{\texttt{obb}: OrientedBoundingBox; \texttt{approach\_offset} (0.08); \texttt{approach\_hint}; \texttt{z\_offset}}
{\texttt{grasp\_pose}, \texttt{pre\_grasp\_pose}: Se3Pose; \texttt{approach\_direction}, \texttt{slide\_axis}: Vec3}
{Front/horizontal grasp for a handle (drawers, doors), with a slide axis.}

\skillcard{geometry\_svc.DepthToPointCloud}
{\texttt{depth}: DepthImage; \texttt{intrinsics}: Matrix3x3}
{PointCloud (camera frame)}
{Back-project a depth image to a camera-frame point cloud.}

\skillcard{geometry\_svc.MaskToWorldPoints}
{\texttt{mask}: Mask; \texttt{depth}: DepthImage; \texttt{intrinsics}: Matrix3x3; \texttt{camera\_pose}: Se3Pose}
{PointCloud (world frame)}
{Back-project a 2D mask to world-frame 3D points.}

\skillcard{geometry\_svc.PixelToWorldPoint}
{\texttt{pixel\_x}, \texttt{pixel\_y}; \texttt{depth}; \texttt{intrinsics}; \texttt{camera\_pose}}
{Vec3 (world point)}
{Back-project a single pixel to a world point.}

\skillcard{geometry\_svc.TransformPoints}
{\texttt{point\_cloud}: PointCloud; \texttt{transform}: Se3Pose}
{PointCloud}
{Apply a rigid transform to a set of points.}

\skillcard{geometry\_svc.BuildWorldConfig}
{\texttt{cameras}: CameraObservation[]; \texttt{object\_masks}: [name, mask, camera\_index]; recon params (\texttt{voxel\_size} 0.005, \texttt{mesh\_alpha} 0.03); \texttt{robot\_joint\_state} (optional); \texttt{target\_obb} (optional)}
{\texttt{config}: WorldConfig, \texttt{mesh\_names}: string[]}
{Reconstruct a named collision-mesh world from camera observations for any
planner.}

\skillcard{geometry\_svc.ComputeDropPosition}
{\texttt{container\_obb}: OrientedBoundingBox; \texttt{clearance} (0.05); \texttt{object\_z\_extent}}
{Vec3 (drop point)}
{Drop position above a container, accounting for clearance and object height.}

\skillcard{geometry\_svc.ComputeXYDistance}
{\texttt{point\_a}, \texttt{point\_b}: Vec3}
{\texttt{distance}: double}
{XY-plane Euclidean distance between two points.}

\skillcard{geometry\_svc.RotateQuatZ90}
{\texttt{quat}: Quaternion (WXYZ)}
{Quaternion}
{Rotate a quaternion 90$^\circ$ about Z (for grasp candidates).}

\paragraph{Robot and environment control.}

\skillcard{robot\_control.GetEEPose}
{\texttt{arm\_id} (0)}
{\texttt{pose}: Se3Pose}
{Current end-effector pose in world frame.}

\skillcard{robot\_control.GoToPose}
{\texttt{pose}: Se3Pose; \texttt{z\_approach}; \texttt{tcp\_offset} ($0,0,-0.1$); \texttt{tolerance} (0.01); \texttt{max\_steps} (120); \texttt{nonblocking}}
{(empty)}
{Move the EE to a target pose via IK (no collision avoidance).}

\skillcard{robot\_control.GoToPoseCartesian}
{\texttt{pose}: Se3Pose; \texttt{lin\_vel\_norm} (1.0); \texttt{ang\_vel\_norm} (2.0); \texttt{timeout\_s} (20)}
{(empty)}
{Move to a pose with smooth Cartesian interpolation.}

\skillcard{robot\_control.ExecuteJointTrajectory}
{\texttt{trajectory}: Trajectory; \texttt{subsample} (1); \texttt{tolerance} (0.01)}
{(empty)}
{Execute a pre-planned joint trajectory (from CuRobo or PyRoKI).}

\skillcard{robot\_control.MoveToJoints}
{\texttt{joint\_config}: JointState; \texttt{tolerance} (0.01); \texttt{max\_steps} (120)}
{(empty)}
{Move directly to a joint configuration.}

\skillcard{robot\_control.GoHome}
{(none)}
{(empty)}
{Move the robot to its safe home configuration.}

\skillcard{robot\_control.GoToPoseArm}
{\texttt{arm\_id}; \texttt{pose}: Se3Pose; \texttt{z\_approach}; \texttt{tcp\_offset}}
{(empty)}
{Multi-arm: move one specified arm to a pose.}

\skillcard{robot\_control.GoToPoseBoth}
{\texttt{pose\_arm0}, \texttt{pose\_arm1}: Se3Pose; \texttt{z\_approach}; \texttt{tcp\_offset}}
{(empty)}
{Multi-arm: move both arms to their poses simultaneously.}

\skillcard{robot\_control.MoveToJointsBoth}
{\texttt{joints\_arm0}, \texttt{joints\_arm1}: JointState; \texttt{tolerance} (0.02); \texttt{max\_steps} (300)}
{(empty)}
{Multi-arm: drive both arms to raw joint targets in lock-step (no IK).}

\skillcard{robot\_control.ApplyPolicyAction}
{\texttt{action}: double[] (e.g. 7-dim $[\Delta x,\Delta y,\Delta z,\Delta r_x,\Delta r_y,\Delta r_z,\text{grip}]$); \texttt{arm\_id} (0)}
{(empty)}
{Apply one low-level policy action to the env controller (VLA rollouts).}

\skillcard{gripper.Open}
{\texttt{arm\_id} (0); \texttt{settle\_steps} (40)}
{GripperState}
{Open the gripper and settle.}

\skillcard{gripper.Close}
{\texttt{arm\_id} (0); \texttt{settle\_steps} (60)}
{GripperState}
{Close the gripper and settle.}

\skillcard{gripper.GetPosition}
{\texttt{arm\_id} (0)}
{GripperState (opening width)}
{Read the current gripper opening.}

\skillcard{gripper.GetPose}
{\texttt{arm\_id} (0)}
{\texttt{pose}: Se3Pose}
{Gripper pose in world frame.}

\skillcard{observation.GetObservation}
{(none)}
{\texttt{cameras}: CameraObservation[] (RGB-D + intrinsics/extrinsics), \texttt{arm\_states}: ArmState[]}
{Full sensor observation from all cameras and arms.}

\skillcard{observation.GetCameraPose}
{\texttt{camera\_name} (e.g. ``agentview'')}
{\texttt{pose}: Se3Pose}
{A named camera's pose in robot base frame.}

\skillcard{sim\_bridge.Init}
{\texttt{env\_name}; \texttt{config\_path} (optional); \texttt{camera\_names}; \texttt{task\_id}; \texttt{suite\_name}}
{\texttt{success}, \texttt{capabilities}: SimCapabilities, \texttt{error\_message}}
{Initialize a simulation environment by name.}

\skillcard{sim\_bridge.Reset}
{\texttt{seed}}
{\texttt{observation}: ObservationResponse}
{Reset the environment for a new episode.}

\skillcard{sim\_bridge.StepOnce}
{\texttt{gripper\_fraction} ($-1$ = no change)}
{\texttt{observation}, \texttt{reward}, \texttt{terminated}, \texttt{truncated}}
{Execute a single low-level MuJoCo timestep.}

\skillcard{sim\_bridge.MoveToJointsBlocking}
{\texttt{target\_joints}: JointState; \texttt{tolerance} (0.01); \texttt{max\_steps} (120)}
{\texttt{observation}, \texttt{reward}, \texttt{terminated}, \texttt{truncated}}
{Move to a joint configuration, blocking until converged.}

\skillcard{sim\_bridge.SetGripper}
{\texttt{fraction} (0 closed $\to$ 1 open); \texttt{arm\_id} (0)}
{(empty)}
{Set the gripper target fraction.}

\skillcard{sim\_bridge.GetState}
{(none)}
{\texttt{arm\_states}: ArmState[], \texttt{cameras}: CameraObservation[], \texttt{objects}: [name, pose] (ground truth)}
{Full sim state, including per-object ground-truth poses.}

\skillcard{sim\_bridge.CheckTaskCompletion}
{(none)}
{\texttt{reward}, \texttt{task\_completed}, \texttt{completion\_rate} (sub-goals satisfied)}
{Query simulator reward and task success.}

\skillcard{sim\_bridge.EnableVideoCapture}
{\texttt{enabled}; \texttt{clear}}
{(empty)}
{Start/stop video-frame capture.}

\skillcard{sim\_bridge.SaveVideo}
{\texttt{output\_path}; \texttt{clear}; \texttt{fps} (20)}
{\texttt{success}, \texttt{file\_path}, \texttt{num\_frames}}
{Encode captured frames to an MP4 and return its path.}

\section{Self-Learning}
\subsection{Pseudo-code}
\begin{algorithm}[H]
\scriptsize 
\DontPrintSemicolon
\caption{Rehearsal-based Graph Optimization}
\label{alg:graph_opt}
\SetKwInOut{KwInput}{In}
\SetKwInOut{KwOutput}{Out}
\noindent\textbf{Input:} $\mathcal{T} = \langle \mathcal{L}, \mathcal{E}, \mathcal{R}, \mathcal{O}, \mathcal{X}, \mathcal{B}, \mathcal{J} \rangle$,  Iterations $M$, Parallel Rollouts $N$\\
\noindent\textbf{Output:} Optimized Task Graph $\mathcal{G}^*$
\BlankLine
\textbf{Initialization:}\;
$\mathcal{G}_0 \leftarrow \mathrm{Graph\_Init}(\mathcal{L})$\;
\quad\tcp*[r]{\tiny Initial graph from language}
$\mathcal{S} \leftarrow \mathrm{Build\_Scene}(\mathcal{G}, \{\mathcal{G}_i\})$\;
\quad\tcp*[r]{\tiny Instantiate sim environment}
\For{$j \leftarrow 1$ \KwTo $M$}{
    \tcp{Step 1: Scene Variational Sampling}
    $\{\hat{s}_i\}_{i=1}^N \sim \mathcal{B}$\;
    \quad\tcp*[r]{\tiny Sample $N$ instances}
    \tcp{Step 2: Parallel Rehearsal}
    \ParallelFor{$i \leftarrow 1$ \KwTo $N$}{
        $\tau_i \leftarrow \mathrm{Rehearsal}(\hat{s}_i, \mathcal{G}_{j-1})$\;
        \quad\tcp*[r]{\tiny Rollout policy}
        $F_i \leftarrow \mathrm{Analyze_\_Failure}(\tau_i, \mathcal{G}, \{\mathcal{G}_i\})$\;
    
    }
    \tcp{Step 3: Graph Refinement}
    $\mathcal{G}_j \leftarrow \mathrm{Graph\_Update}(\{F_1, \dots, F_N\})$\;
    \quad\tcp*[r]{\tiny Optimize via LLM}
}
\Return $\mathcal{G}^* \leftarrow \mathcal{G}_M$\;
\end{algorithm}

\subsection{Sample Feedback}
Below is a sample feedback from running pan of 30 concurrent environment. At each phase, the difference between simulated object states and contacts. The differences  before and after each phase is recorded and provided to LLM agent to iteratively improve the graph. For example, if the oriented bounding box of pan is not overlapped with the oriented bounding box of the stove, and they are not in contact, the LLM can reason and infer as a failure.

\hypersetup{
    colorlinks=true,
    linkcolor=blue!55!black,
    urlcolor=blue!55!black,
    citecolor=blue!55!black
}

\tcbset{
  rolloutsummary/.style={
    enhanced,
    breakable,
    colback=gray!2,
    colframe=black!25,
    boxrule=0.35pt,
    arc=1.5pt,
    left=6pt,
    right=6pt,
    top=5pt,
    bottom=5pt,
    before skip=6pt,
    after skip=6pt,
  },
  rolloutfailure/.style={
    enhanced,
    breakable,
    colback=gray!1,
    colframe=black!22,
    boxrule=0.35pt,
    arc=1.5pt,
    left=6pt,
    right=6pt,
    top=5pt,
    bottom=5pt,
    before skip=6pt,
    after skip=6pt,
  },
  rolloutnote/.style={
    enhanced,
    breakable,
    colback=blue!2,
    colframe=blue!18,
    boxrule=0.35pt,
    arc=1.5pt,
    left=6pt,
    right=6pt,
    top=5pt,
    bottom=5pt,
    before skip=6pt,
    after skip=6pt,
  }
}

\newcommand{\nodeid}[1]{\texttt{#1}}
\newcommand{\metric}[1]{\textbf{#1}}
\newcommand{\failtag}[1]{\texttt{\textbf{#1}}}
\newcommand{\code}[1]{\texttt{#1}}

\begin{tcolorbox}[rolloutnote]
\small
\textbf{Summary.}
In the first rehearsal iteration, perception succeeds across all sampled environments, while downstream failures arise from grasp instability and incorrect pan placement. The placement node fails because the pan is released away from the burner footprint, yielding zero burner coverage.
\end{tcolorbox}

\begin{tcolorbox}[rolloutsummary]
\small
\textbf{Iteration 1 Feedback} \hfill \textbf{Terminal coverage: 0.00}

\vspace{4pt}
\renewcommand{\arraystretch}{1.18}
\begin{tabularx}{\linewidth}{@{}p{0.23\linewidth}p{0.18\linewidth}X@{}}
\toprule
\textbf{Node} & \textbf{Validation} & \textbf{Feedback} \\
\midrule
\nodeid{perceive\_pan}
& \metric{1.00 (30/30)}
& Perception publishes the target OBB for the pan handle. \\

\nodeid{perceive\_burner}
& \metric{1.00 (30/30)}
& Perception publishes the container OBB for the burner. \\

\nodeid{grasp\_pan}
& \metric{0.83 (25/30)}
& The pan is considered grasped when it remains in contact with a robot link at the exit of the grasp subgraph. \\

\nodeid{place\_pan}
& \metric{0.57 (17/30)}
& The pan should be released by the gripper and cover at least 70\% of the burner XY footprint. All evaluated placement attempts fail because the pan does not cover the burner footprint. \\
\bottomrule
\end{tabularx}
\end{tcolorbox}

\begin{tcolorbox}[rolloutfailure]
\small
\textbf{Representative Failure Cases}

\vspace{2pt}
\begin{itemize}[leftmargin=1.2em, itemsep=3pt, topsep=2pt]
    \item \failtag{env 3, grasp\_pan}: grasp failed with the gripper still open.
    \[
    \code{gripper\_open\_fraction}=1.0,\quad
    \code{pan\_z}=0.024,\quad
    \code{contact}=[\code{table\_top}]
    \]
    The subgraph returned \code{failed}; only \code{ee\_pose\_at\_grasp} was bound. 

    \item \failtag{env 9, grasp\_pan}: grasp failed after gripper closure.
    \[
    \code{gripper\_open\_fraction}: 1.0 \rightarrow 0.0,\quad
    \code{pan\_z}=0.024,\quad
    \code{contact}=[\code{table\_top}]
    \]
    The robot closed the gripper, but the pan remained on the table and the node returned \code{failed}. The elapsed time was \code{4.377 s}.

    \item \failtag{place\_pan}: the evaluated placement failure trials has:
    \[
\code{pan\_coverage\_of\_burner}\leq 0.7
    \]
    Although the pan was released from the gripper, its projected bottom footprint did not sufficiently overlap with the burner footprint.
\end{itemize}
\end{tcolorbox}

\begin{tcolorbox}[rolloutnote]
\small
\textbf{Optimization signal.}
The feedback suggests that the next graph update should revise the placement target and release pose for \nodeid{place\_pan}, while also improving grasp robustness for the failed \nodeid{grasp\_pan} environments.
\end{tcolorbox}

\section{Sample LLM Prompts and Outputs}

\subsection{Behavior Agent Prompt}
\begin{lstlisting}
You are a robotics task planner. Given a manipulation task description,
you produce **only the workflow topology** — the top-level node graph
(subgraph nodes + end nodes), the edges and conditional_edges connecting
them, and each subgraph's inputs/outputs/exit-values/skill choice. You do
**not** generate internal nodes/edges of any subgraph. The universal
`subgraph_agent` is invoked separately per subgraph to fill in its inner
state machine.

## Output format

A single ` ```python` fenced block (no file path) that builds the workflow
scaffold with `vos.builder.WorkflowSpec` and binds it to a module-level
variable named ``spec``:

```python
from vos.builder import WorkflowSpec, START
from vos.runtime.workflow import ServiceCall

spec = WorkflowSpec(name="<task>", description="<task prompt>")

# 1. Declare every subgraph (metadata only — nodes/edges get filled in later).
spec.declare_subgraph(
    "<sg_def_name>",                  # e.g. "target_sg"
    skill="<MORSL skill name from the Available Skills list>",
    description="<natural-language goal for this subgraph instance>",
    inputs=,
    outputs=,
    exit_success_values=["<success_exit>"],
    on_error="<failure_exit>",
    stage="<grasp|transport|place>",  # OPTIONAL; see "Stage tag" below.
)

# 2. Place the top-level subgraph nodes that reference those declarations.
spec.add_subgraph_node("<sg_node_name>", ref="<sg_def_name>")

# 3. Add end nodes — typically one success ("done") and one failure ("abort").
spec.add_end("done", status="success")
spec.add_end(
    "abort",
    status="failure",
    recovery=[
        ServiceCall("gripper.v1.Gripper", "Open", {}),
        ServiceCall("robot_control.v1.RobotControl", "GoHome", {}),
    ],
)

# 4. Wire the entry edge and conditional dispatch per subgraph node.
spec.add_edge(START, "<entry_subgraph_node>")
spec.add_conditional_edges(
    "<sg_node_name>",
    {"<exit_value>": "<next_node>", ...},
    router_field="exit",
)
```

The pipeline imports `vos.builder` (already available; do not pip
install), executes the block in a sandbox, picks up the module-level
`spec` variable, serializes it to a workflow-spec dict, and forwards it
to the per-subgraph generators.

Use a 1:1 naming convention between the outer node name (passed to
`add_subgraph_node`) and the inner subgraph def name (passed to
`declare_subgraph`), or use the same name for both — both work.

### Hard rules for the Python block

1. The block MUST end with a module-level variable named ``spec`` bound
   to a ``WorkflowSpec`` instance.
2. Imports: ``from vos.builder import WorkflowSpec, START`` and
   ``from vos.runtime.workflow import ServiceCall`` are pre-provided in
   the sandbox.
3. No I/O, no other imports.
4. Subgraph `inputs` / `outputs` MUST be dicts of `{name: proto_type_str}`.
   Each value is a **bare proto type string** (e.g. ``"OrientedBoundingBox"``,
   ``"Mask"``, ``"PointCloud"``) — never a ``Ref`` and never a nested dict.
   Cross-subgraph data flow is established implicitly by `add_edge` /
   `add_conditional_edges` plus matching input/output **names** between
   subgraphs — you do NOT wire it here.
5. Every `declare_subgraph(...)` call MUST pass BOTH `exit_success_values=`
   AND `on_error=` as keyword arguments — they are required and have no
   defaults. Omitting either raises
   `declare_subgraph() missing 2 required keyword-only arguments` and
   the whole spec rejects. Conventional values: `exit_success_values=["done"]`
   and `on_error="abort"`, paired with `spec.add_end("done", status="success")`
   and `spec.add_end("abort", status="failure", recovery=[...])`.

### Stage tag (optional but recommended)

Set ``stage="grasp"``, ``"transport"``, or ``"place"`` on every
``declare_subgraph`` that participates in the pick-and-place pipeline.
The iter-1 mechanical-swap engine
(``vos/refine/iter1_swap.py``) groups per-env checkpoint outcomes by
``stage`` to compute per-stage pass-rates and decide which canonical
swap (graspgen↔OBB, free-space↔collision-aware transport, subpart↔parent
drop anchor) to apply for the next iter.

- ``grasp`` — any subgraph whose exit is "object held in gripper"
  (skills ``grasp_moe``, ``grasp_multi``, ``grasp_curobo_obb``).
- ``transport`` — moves the held object from grasp pose to above the
  drop zone (``transport_to_drop`` when modeled as its own subgraph).
- ``place`` — the release leg: compute drop pose + drop-offset
  correction + descend/release (``transport_to_drop`` when modeled
  end-to-end, or a dedicated placement subgraph).

When a subgraph does not belong to the canonical taxonomy (e.g. a
perception or pre-grasp staging subgraph), omit ``stage``. The
iter-1 engine skips unstaged subgraphs during stage rollups; explicit
``None`` is fine.

## Hard rules

1. **Edges and conditional_edges.** For each subgraph node, every one of
   its `exit_success_values` PLUS its `on_error` symbol must appear as a
   key in the corresponding `add_conditional_edges` mapping, and every
   mapping target must be a node declared at the top level. Every end
   node must be reachable from `START`.
2. **The entry node must be a subgraph node** (not an end node).
3. **No internal nodes / edges / on_error** for any subgraph in your
   output. The universal subgraph_agent fills those in per subgraph.
4. **Inputs and outputs.** A subgraph's declared `inputs` must equal its
   skill's `required_inputs` after `<name>` substitution, and every
   input must have an upstream subgraph on some path to it that produces
   a matching output name with a matching proto type. Declared `outputs`
   must be a subset of the skill's `produces_outputs` (omit outputs
   nothing downstream consumes).
5. **Pick the right specialized variant.** When multiple variants of a
   role appear in Available Skills (e.g. `perception_multi` vs
   `perception_single`, `grasp_curobo_obb` vs `grasp_direct_ik`), read
   each skill's *When to use* guidance and pick the best fit — default
   to the more robust / collision-aware variant when both are listed.
   Skills whose required services are not deployed do not appear in
   Available Skills.

## Discovery via tools

You may call:

- `read_skill_reference(skill_name, doc_name)` to load the long-form
  rationale for a skill (the references listed in the skill's frontmatter).
- `read_skill_example(skill_name, example_name)` to load a sample
  subgraph the per-skill subgraph_agent will start from.
- `report_missing_capability(name, why)` if no skill in the catalog
  covers a step the task requires. The build aborts with a structured
  report.

Use these sparingly — the always-loaded catalog already shows skill
descriptions, tags, exit_conditions, and produces_outputs.

# Workflow v3 spec — shared core

This is the structural spec every codegen subagent shares. It defines the
top-level workflow shape, the SubgraphDef shape, node types, edge
semantics, `$ref` syntax, and validation rules.

## Top-level workflow

```json
{
  "version": 3,
  "meta": { "name": "...", "description": "..." },
  "nodes": {
    "<sg_node_name>": { "type": "subgraph", "ref": "<sg_def_name>" },
    "done":  { "type": "end", "status": "success" },
    "abort": { "type": "end", "status": "failure",
               "recovery": [ { "service": "...", "method": "...", "inputs": {} } ] }
  },
  "edges": [["START", "<first_subgraph_node>"]],
  "conditional_edges": {
    "<sg_node_name>": {
      "router_field": "exit",
      "mapping": { "<exit_value>": "<next_node>", ... }
    }
  },
  "subgraphs": { "<sg_def_name>": { /* SubgraphDef */ } }
}
```

`START` and `END` are virtual node names. The top level orchestrates
subgraph nodes and end nodes; cross-subgraph routing is via the
`conditional_edges` block reading each subgraph's `exit` value.

`meta` is an open string→string map; recognized keys include `name`,
`description`, `runtime` (`direct_real`), `observation_stream_hz`, and
`validate_checkpoints` (`"true"` opts the executor into enforcing each
subgraph's `validate=True` postcondition checkpoints at real-execution
time — **currently a no-op**; reserved seam).

## SubgraphDef shape

```json
{
  "skill":  "<MORSL skill name; echoed from your spec>",
  "inputs":  { "<name>": "<proto.type.string>" },
  "outputs": { "<name>": { "$ref": "<node>.<field>" } },
  "nodes": {
    "<node_name>": { /* NodeDef */ },
    "<terminal_marker>": { "type": "noop" }
  },
  "edges": [
    ["START", "<first_node>"],
    ["<first_node>", "<second_node>"],
    ["<terminal_marker>", "END"]
  ],
  "conditional_edges": { },
  "exit": { "router_field": null, "success_values": ["<success_exit>"] },
  "on_error": "<failure_exit_value>"
}
```

Scripts go in separate fenced blocks, namespaced under
`scripts/<subgraph_name>/`:

````python:scripts/<subgraph_name>/<script>.py
...
````

## Node types

The two production dispatch types are `tool` and `script`. `noop` and
`router` are control-flow markers. `subgraph` and `end` only appear at
the top level of the workflow (not inside subgraphs).

```json
{ "type": "tool", "tool": "gripper.Open",
  "inputs": { "settle_steps": 40 } }                       /* gRPC method */

{ "type": "tool", "tool": "sam3.SegmentBox",
  "inputs": { "image": { "$ref": "obs.rgb" }, "box": { "$ref": "perceive.box" } } }

{ "type": "tool", "tool": "run_policy",
  "inputs": { "observation_stream": { "$ref": "in.observation_stream" } } }  /* atomic MORSL skill */

{ "type": "tool", "tool": "libero_pi05",
  "inputs": { "prompt": "...", "max_windows": 25 } }       /* learned policy */

{ "type": "tool", "tool": "track_object", "streaming": true,
  "inputs": { "observation_stream": { "$ref": "in.observation_stream" } } }

{ "type": "script", "script": "scripts/<subgraph_name>/foo.py",
  "inputs": { ... } }                                      /* canonical bundle script
                                                              or rare inline helper */

{ "type": "noop" }                                          /* named terminal marker */
{ "type": "router", "script": "scripts/<sg>/route.py", "inputs": { ... } }
```

- **`tool`** — the canonical dispatch for **anything**: gRPC service
  methods (auto-registered as `<package>.<MethodName>`, e.g.
  `observation.GetObservation`, `gripper.Open`,
  `geometry_svc.FilterAndComputeOBB`), MORSL skills (registered by skill
  name, e.g. `run_policy`, `track_object`), and learned policies
  (registered by policy name). `tool:` is always a flat name; never write
  `<package>.v1.<Service>` or include a separate `method:` field. The
  runtime resolves the proto FQN internally.
- **`script`** — local Python file in the workflow folder. Prefer to
  point at a canonical bundle script (listed in the chosen skill's
  "Canonical scripts" table); only emit your own inline Python for
  ad-hoc helpers that no canonical and no tool covers.
- **`noop`** — empty body. Used as a named subgraph terminal so the node
  name becomes the subgraph's exit value when `exit.router_field=null`.
- **`router`** — Send dispatch. Function returns either a string target
  for static routing or a list of `{"to": "...", "inputs": {...}}` dicts
  for dynamic fan-out.

*Legacy node types (`type: service`, `type: skill`, `type: policy`)
have been retired. The validator rejects them with a precise migration
message. Always emit `type: tool` with the flat name instead.*

`streaming: true` is only valid on `tool` and `script` nodes. A
streaming node has **no outgoing edges** — it's a pure source. Consumers
read its latest published value via `{"$ref": "<node>"}`. The chosen
tool's bundle contract must declare `contract.streaming: true`.

## Edge semantics

- **Static edges**: `["src", "dst"]` pairs.
- **Multiple outgoing edges from one node = parallel super-step.**
- **`conditional_edges`** dispatches on a router field. From a
  non-router source the field is on the source's output (set
  `router_field` to its name). From a router source set
  `router_field: null`.
- **`exit.router_field: null`** means the subgraph's exit value is the
  name of the terminal node (the one whose edge points to `END`). For
  this to work, declare your terminals as `noop` nodes named after the
  exit values (e.g. `"found": {"type": "noop"}` with edge
  `["found", "END"]`).
- **`exit.router_field: "<field>"`** reads the field on the terminal
  node's output as the exit value.
- **`on_error`** is an optional subgraph-level catch: when any node
  raises, the runtime emits this string as the subgraph's exit value
  (bypassing the terminal-node read). The `on_error` symbol is the
  subgraph's single failure exit. It is **never** a declared node and
  **never** a `conditional_edges` mapping target — failures surface
  only by raising.

### `perception_multi`

Molmo point-prompt + SAM3 point/text segmentation + depth-to-3D fusion, ending in a FilterAndComputeOBB to produce a clean OBB and mask. Single path; molmo-only perception (no DINO/VLM).

### `perception_single`

Fast single-path object detection: DINO broad detect + VLM disambiguation + SAM3 segmentation + depth-to-3D fusion, ending in a FilterAndComputeOBB to produce a clean OBB and mask. Best for uncluttered scenes with visually distinct targets, or platforms where only DINO/VLM/SAM are deployed.

...

## Available Tools (flat catalog)

| Tool | Transport | Tags | Summary |
|------|-----------|------|---------|
| `curobo.PlanDirectedLinear` | grpc | curobo | CuRobo.PlanDirectedLinear :: PlanDirectedLinearRequest → PlanDirectedLinearResponse |
| `curobo.PlanGraspMotion` | grpc | curobo | CuRobo.PlanGraspMotion :: PlanGraspMotionRequest → PlanGraspMotionResponse |
| `curobo.PlanLinear` | grpc | curobo | CuRobo.PlanLinear :: CuRoboPlanLinearRequest → CuRoboPlanLinearResponse |
| `curobo.PlanToGraspPoses` | grpc | curobo | CuRobo.PlanToGraspPoses :: CuRoboPlanGraspRequest → CuRoboPlanGraspResponse |
| `curobo.PlanToPose` | grpc | curobo | CuRobo.PlanToPose :: CuRoboPlanToPoseRequest → CuRoboPlanToPoseResponse |
| `curobo.PlanWithGraspedObject` | grpc | curobo | CuRobo.PlanWithGraspedObject :: CuRoboPlanGraspedRequest → CuRoboPlanGraspedResponse |
...

## Task

Pick up the object on the table and place it at the designated target location. Perceive the scene to locate the object, grasp it, transport it above the target, and release it so it rests stably at the target.

\end{lstlisting}

\subsection{Behavior Agent Output}
\begin{lstlisting}
```python
from vos.builder import WorkflowSpec, START
from vos.runtime.workflow import ServiceCall

spec = WorkflowSpec(name="pick_and_place", description="Pick up the object on the table and place it at the designated target location.")

# 1. Declare subgraphs
spec.declare_subgraph(
    "target_sg",
    skill="perception_single",
    description="Locate the object on the table",
    inputs={},
    outputs={
        "target_obb": "OrientedBoundingBox",
        "target_mask": "Mask",
        "target_cloud": "PointCloud",
    },
    exit_success_values=["found"],
    on_error="abort",
)

spec.declare_subgraph(
    "container_sg",
    skill="perception_single",
    description="Locate the target location",
    inputs={},
    outputs={
        "container_obb": "OrientedBoundingBox",
        "container_mask": "Mask",
        "container_cloud": "PointCloud",
    },
    exit_success_values=["found"],
    on_error="abort",
)

spec.declare_subgraph(
    "grasp_sg",
    skill="grasp_multi",
    description="Grasp the object on the table",
    inputs={
        "target_obb": "OrientedBoundingBox",
        "target_mask": "Mask",
        "target_cloud": "PointCloud",
    },
    outputs={"ee_pose_at_grasp": "Se3Pose"},
    exit_success_values=["grasped"],
    on_error="abort",
    stage="grasp",
)

spec.declare_subgraph(
    "transport_sg",
    skill="transport_to_drop",
    description="Transport the object to the target location",
    inputs={
        "target_obb": "OrientedBoundingBox",
        "target_mask": "Mask",
        "container_obb": "OrientedBoundingBox",
        "container_mask": "Mask",
        "ee_pose_at_grasp": "Se3Pose",
    },
    outputs={},
    exit_success_values=["placed"],
    on_error="abort",
    stage="place",
)

# 2. Place subgraph nodes
spec.add_subgraph_node("target_sg", ref="target_sg")
spec.add_subgraph_node("container_sg", ref="container_sg")
spec.add_subgraph_node("grasp_sg", ref="grasp_sg")
spec.add_subgraph_node("transport_sg", ref="transport_sg")

# 3. Add end nodes
spec.add_end("done", status="success")
spec.add_end(
    "abort",
    status="failure",
    recovery=[
        ServiceCall("gripper.v1.Gripper", "Open", {}),
        ServiceCall("robot_control.v1.RobotControl", "GoHome", {}),
    ],
)

# 4. Wire edges
spec.add_edge(START, "target_sg")
spec.add_conditional_edges(
    "target_sg",
    {"found": "container_sg", "not_found": "abort"},
    router_field="exit",
)
spec.add_conditional_edges(
    "container_sg",
    {"found": "grasp_sg", "not_found": "abort"},
    router_field="exit",
)
spec.add_conditional_edges(
    "grasp_sg",
    {"grasped": "transport_sg", "failed": "abort"},
    router_field="exit",
)
spec.add_conditional_edges(
    "transport_sg",
    {"placed": "done", "blocked": "abort"},
    router_field="exit",
)
```
\end{lstlisting}
\subsection{Skill Agent Prompt}

\begin{lstlisting}
# Subgraph agent

You generate **one subgraph** for a v3 robotics workflow. The
coordinator chose your input spec (subgraph name, declared
inputs/outputs/exit-values, and the MORSL skill this subgraph should
use). Your output is a Python script that builds the subgraph with
the `vos.builder` library, plus any inline Python scripts referenced
by `type="script"` nodes.

## Your context window contains

1. The shared workflow spec (top-level shape, node types, edge
   semantics, `$ref` syntax, validation rules) — see
   `_workflow_spec.md`.
2. The chosen skill's SKILL.md body (this is the per-skill guidance —
   recommended node sequence, hard rules, exit-value semantics, and
   contract — including whether the skill is `streaming: true`).
3. The filtered tool catalog: only the tools listed in the skill's
   `allowed-tools` frontmatter, intersected with what is deployed.
4. The chosen skill's `canonical_scripts` list (for composite skills) —
   file references the subgraph may use as `type="script"` nodes.
5. The coordinator-supplied subgraph spec (name, inputs, outputs,
   exit values, context).

## Your job

Compose the minimal sequence of nodes and edges that:

1. Consumes declared inputs (referenced as `Ref(f"in.{name}")`).
2. Produces values bound to the declared outputs via `sg.set_outputs(...)`.
3. Names every success-path exit via `sg.add_exit(name)` (creates the
   terminal `noop` marker), and names the single failure-path exit via
   `sg.set_on_error(value)`.
4. Reaches each success-exit marker on its own path with an explicit
   edge to `END`.
5. Calls only tools in the filtered catalog and scripts in the skill's
   `canonical_scripts` list. If you need a primitive that's missing,
   call `report_missing_capability(name, why)`.

Postcondition checkpoints (`sg.add_checkpoint(...)`) are authored by a
separate `checkpoint_agent` in a follow-on pass. **Do not** declare
any checkpoints yourself — emit structure (nodes, edges,
`set_outputs(...)`, `set_on_error(...)`) only.

### Exit-value rule (HARD, single rule)

There is **one and only one** namespace collision question, and it's
answered by which field the exit value lives in:

| Form | Is it a node? | Where in code |
|---|---|---|
| Success exit (default `set_exit_router` is unset) | **YES** — call `sg.add_exit(name)` (creates a `noop` and edges to `END` are your responsibility) | `sg.add_exit("found")` + `sg.add_edge("found", END)` |
| Success exit when `set_exit_router(router_field=..., success_values=[...])` is used | **NO** — string field-values returned by the terminal node, never node names | `sg.set_exit_router(router_field="exit", success_values=["ok"])` |
| `on_error` | **NO** — single failure symbol; never declare a node with that name | `sg.set_on_error("not_found")` |

Wrong patterns the validator rejects (with rule IDs):

- **S9**: `sg.set_on_error("failed")` plus `sg.add_node("failed", type="noop")` — the `failed` node is forbidden.
- **S10**: routing `"false" -> "failed"` in a `sg.add_conditional_edges(...)` mapping — `failed` cannot be a conditional-edge target. Failure is surfaced by raising, not routing.
- **S11**: `sg.set_exit_router(..., success_values=["grasped"])` then no terminal node returns `grasped` in its `exit` field — every success value must be producible.

### Conditional-edge rules (HARD)

**Source rule:** the first argument to `sg.add_conditional_edges(src, ...)` is the source node — that node must be a real producer of the field named in `router_field`. Terminal `noop` markers (created via `add_exit`) are NOT routers; they have a single outgoing edge to `END` and must NOT be a source.

**Target rule:** every value in a `mapping` must be a **declared node name** (or `START` / `END`) AND must not equal `on_error`. The single failure exit lives only via `set_on_error` and is reached by raising, never by routing.

**How to express "this subgraph's postcondition must hold"** (e.g. "the gripper is actually holding the target after `close`"): do NOT add a node that re-checks the end state and raises, and do NOT route to `on_error`. The follow-on `checkpoint_agent` will declare the postcondition via `sg.add_checkpoint(...)` against privileged state.

**Mid-subgraph preconditions** (a check whose failure means *subsequent nodes in this subgraph cannot run*) are different — for those, insert a `type="script"` guard that raises; the raise propagates to `on_error`:

```python
# scripts/<sg>/require_cloud.py
def run(ctx, found: bool) -> None:
    if not found:
        raise RuntimeError("target not detected; cannot plan a grasp")
    return None
```
```python
sg.add_node("require_cloud", type="script", script="scripts/<sg>/require_cloud.py",
            inputs={"found": Ref("perceive.found")})
sg.add_edge("perceive", "require_cloud")
sg.add_edge("require_cloud", "compute_grasp")
sg.set_on_error("not_found")
```

Use a raising guard only for genuine mid-flow preconditions. For *postconditions* — the end-state promise of the subgraph — leave them to the `checkpoint_agent` follow-on pass; do not declare them here.

If the chosen skill's contract has `streaming: true`, the node invoking it must declare `streaming=True` and have **no outgoing edges** — it's a pure source. Other nodes consume its latest snapshot via `Ref("<this_node_name>")`. A streaming node must still be declared as an edge target from `START` (or another super-step source) so the runtime spawns it.

If an ad-hoc Python step is needed that no canonical script covers, call `request_inline_script(name, signature, purpose, body_hint)` — the coder subagent will emit the script and return its path. Reference the returned path in a `type="script"` node.

## Output format

A single ` ```python` fenced block (no file path) that builds the
subgraph by assigning to a module-level variable named ``sg``:

```python
from vos.builder import Subgraph, Ref, START, END

sg = Subgraph(name="<the subgraph name from your spec>", skill="<the chosen skill name>")

# Declare cross-subgraph inputs (from your spec):
sg.add_input("target_obb", proto_type="OrientedBoundingBox")

# Add the nodes that make up the state machine:
sg.add_node("observe", type="tool", tool="observation.GetObservation")
sg.add_node("perceive", type="script", script="scripts/<sg>/perceive.py",
            inputs={"cameras": Ref("observe.cameras"), "object": "{{target_full}}"})

# Declare success markers — these create `noop` nodes whose names equal
# the exit value. They MUST appear in the edge list with an edge to END.
sg.add_exit("found")

# Wire the edges.
sg.add_edge(START, "observe")
sg.add_edge("observe", "perceive")
sg.add_edge("perceive", "found")
sg.add_edge("found", END)

# Bind subgraph-level outputs declared by your spec, if any.
sg.set_outputs(target_obb=Ref("perceive.obb"), target_mask=Ref("perceive.mask"))

# Declare the failure-path exit symbol.
sg.set_on_error("not_found")

# Do NOT call sg.add_checkpoint(...) — the checkpoint_agent runs after
# you and authors all postconditions for the whole workflow at once.
```

The pipeline imports `vos.builder` (already available; do not pip
install), executes the block in a sandbox, picks up the module-level
`sg` variable, runs the v3 structural validator (S1–S11) against it,
and serializes to JSON.

### Hard rules for the Python block

1. The block MUST end with a module-level variable named ``sg`` bound to
   a ``Subgraph`` instance. Anything else (including stray top-level
   ``print`` calls or ``Workflow`` instances) is rejected.
2. Imports: ``from vos.builder import Subgraph, Ref, START, END`` is
   provided in the sandbox — you may re-import it (idempotent) but no
   other imports are needed.
3. No I/O: do not open files, call ``requests``, spawn threads, or
   import packages beyond ``vos.builder``. Imports outside the allow
   list are rejected.
4. No mutation of nodes/edges after they're added (the builder has no
   `remove`/`rename`/`replace` — re-author from scratch instead).

### Inline-script blocks (unchanged)

Optional ` ```python:scripts/<sg>/<file>.py` blocks for inline scripts —
**only for paths whose stem is NOT in the chosen skill's "Canonical
scripts" table above**. If a `type="script"` node points to
`scripts/<sg>/foo.py` and `foo` is a canonical-script stem, the bundle's
canonical implementation is materialized into the workflow directory
automatically; emitting your own ``` ```python:scripts/<sg>/foo.py``` ```
block **overrides** the canonical with whatever Python you write, which
is the leading cause of correctness regressions in this pipeline
(wrong proto imports, simplified math, dropped parameters).
Re-emit a canonical-stem script only after calling
`request_inline_script` to get explicit approval — and prefer to leave
the canonical alone.

Inline-script blocks are distinguished from the subgraph-builder block
by their fence info: ``` ```python:scripts/... ``` (with a path) is an
inline script, ``` ```python ``` (no path) is the subgraph builder.

## Patterns

**Linear pipeline** (most common): only the success exit is a node;
the failure exit lives in `on_error` and is NOT a node.

```python
sg.add_node("a", type="tool", tool="...")
sg.add_node("b", type="script", script="scripts/<sg>/b.py", inputs={...})
sg.add_node("c", type="tool", tool="...")
sg.add_exit("found")
for u, v in [(START, "a"), ("a", "b"), ("b", "c"), ("c", "found"), ("found", END)]:
    sg.add_edge(u, v)
sg.set_on_error("not_found")
```

**Streaming side-car** (when one of your nodes is a streaming-skill
producer that other nodes need to read continuously):

```python
sg.add_node("tracker", type="tool", tool="track_object", streaming=True,
            inputs={...})
sg.add_node("consumer", type="tool", tool="...",
            inputs={"pose": Ref("tracker")})
sg.add_exit("done")
sg.add_edge(START, "tracker")    # spawned; never blocks downstream
sg.add_edge(START, "consumer")
sg.add_edge("consumer", "done")
sg.add_edge("done", END)
sg.set_on_error("failed")
```

**Conditional branch** (router_field on a non-router source). Both
mapping targets must be **declared nodes** — never `on_error`:

```python
sg.add_node("branch", type="tool", tool="...", inputs={...})   # emits a routing field
sg.add_node("retry_step", type="tool", tool="...", inputs={...})
sg.add_exit("alt_done")
sg.add_exit("done")
sg.add_conditional_edges("branch",
    {"true": "done", "false": "retry_step"}, router_field="ok")
sg.add_edge("retry_step", "alt_done")
sg.add_edge("done", END)
sg.add_edge("alt_done", END)
sg.set_on_error("failed")
```

If the false branch should bail to the failure exit, raise instead
(see the raising-guard pattern above) — do NOT make `on_error` a mapping
target. (Postconditions go in `add_checkpoint`, not in a routing branch
or a node that re-checks-and-raises.)

**Send (dynamic fan-out)**: declare a `type="router"` node whose script
returns `[{"to": "process", "inputs": {"item": x}}, ...]`. Each Send
spawns one copy; outputs collect as a list under the router's name.

## What changes vs. legacy

In the legacy system there was a separate Python class per subagent
(perception_multi, grasp_curobo, motion, …). In MORSL, **you are
universal**: the per-skill node-flow guidance comes from the chosen
skill's SKILL.md, not from a domain-specific subagent class. Read the
SKILL.md body (it's in your context) — that's where the recommended
node flow, hard rules, and exit-value mappings live.

## Discovery tools

- `read_skill_reference(skill_name, doc_name)` — load a long-form
  reference doc bundled with the skill. Use this when the SKILL.md
  body's "See also" links the doc and you want the deeper rationale.
- `read_skill_example(skill_name, example_name)` — load a sample
  subgraph the bundle ships, useful as a starting point. (Examples
  may still be in JSON form; convert them to `vos.builder` calls when
  you reuse them.)
- `report_missing_capability(name, why)` — flag a gap; the build
  surfaces it instead of producing broken output.
- `request_inline_script(name, signature, purpose, body_hint)` —
  delegate ad-hoc Python to the coder subagent.

## Postconditions

Every subgraph MUST declare **≥ 1 `validate=True` checkpoint** (with a
non-empty `rationale`) via
`sg.add_checkpoint(name, predicate, *, diagnostics=None, rationale="",
validate=True, weight=1.0)`. **The parser rejects a subgraph with zero
`validate=True` checkpoints and re-prompts you.** Keep the total ≤ 6.
The rehearsal harness evaluates each predicate against a privileged-state
`World` snapshot at subgraph exit and feeds the results back into the
next iter's prompt — without checkpoints the refine loop has no
localized signal and cannot improve anything beyond the binary terminal
verdict.

Granularity is **subgraph-or-coarser**. A typical subgraph declares
ONE `validate=True` checkpoint (the postcondition the subgraph promised
to satisfy — e.g. `target_held`, `ee_above_target`, `target_in_container`)
plus 0-3 `validate=False` probes (`object_lifted`, `gripper_open_fraction`,
etc.) that add diagnostic richness without blocking downstream evaluation.

A `validate=True` checkpoint is also **where runtime postcondition
verification lives**: never add a node that re-checks the subgraph's end
state and raises — declare the postcondition as a `validate=True`
checkpoint on the subgraph that produced the state.

Anchor predicates to privileged quantities only — positions, contacts,
joint state, AABBs, cavity bounds. NEVER reference camera-derived
features (mask IoU, detection confidence); those are what we're trying
to validate, so using them in the predicate makes the checkpoint
circular. Use **scene-spec object ids** as literal name strings — the
`id` from `scene_spec.json` (e.g. `w.body("alphabet soup")`) is the
canonical name that flows through perception inputs, the rehearsal
`World`, and these checkpoint predicates. Never use the original noun
phrase, a free abbreviation, `{{...}}`, or `Ref(...)`.

**Important**: the `description` field of your subgraph spec may use a
paraphrased name (the coordinator should match scene-spec ids verbatim,
but treat that as best-effort context). The scene-spec ids list in your
context is the ground truth for body names. When you write a
`w.body("...")` literal in a checkpoint predicate, **pick the exact spec
id from the scene-spec context, not the word from the description**.
For example, if the description says *"Place the soup can in the basket"*
but the spec ids are `{"alphabet soup", "basket"}`, the predicate MUST
be `w.body("alphabet soup").is_in(w.body("basket"))` — using `"soup can"`
would raise `BodyNotFoundError` at runtime and the checkpoint would
silently fail.

If the chosen skill's SKILL.md has a `## Checkpoints` section, start
from the canonical checkpoint(s) it lists for that skill kind. For the
full reference — `Checkpoint` semantics, the `World` API surface
(`world.body(...)`, `body.is_grasped()`, `body.is_in(...)`, temporal
helpers, history), `validate`-vs-probe rules, and worked predicate
examples for each subgraph kind — see the loaded `_checkpoints_api.md`
include.

The checkpoints are emitted as part of the same Python builder block
that builds the subgraph; they are NOT serialized into `workflow.json`
and they do NOT change the state machine. The pipeline writes them to a
sidecar `<workflow_dir>/checkpoints/<sg>.py` module that the rehearsal
harness loads at refine time.

# Workflow v3 spec — shared core

This is the structural spec every codegen subagent shares. It defines the
top-level workflow shape, the SubgraphDef shape, node types, edge
semantics, `$ref` syntax, and validation rules.

## Top-level workflow

```json
{
  "version": 3,
  "meta": { "name": "...", "description": "..." },
  "nodes": {
    "<sg_node_name>": { "type": "subgraph", "ref": "<sg_def_name>" },
    "done":  { "type": "end", "status": "success" },
    "abort": { "type": "end", "status": "failure",
               "recovery": [ { "service": "...", "method": "...", "inputs": {} } ] }
  },
  "edges": [["START", "<first_subgraph_node>"]],
  "conditional_edges": {
    "<sg_node_name>": {
      "router_field": "exit",
      "mapping": { "<exit_value>": "<next_node>", ... }
    }
  },
  "subgraphs": { "<sg_def_name>": { /* SubgraphDef */ } }
}
```

`START` and `END` are virtual node names. The top level orchestrates
subgraph nodes and end nodes; cross-subgraph routing is via the
`conditional_edges` block reading each subgraph's `exit` value.

`meta` is an open string→string map; recognized keys include `name`,
`description`, `runtime` (`direct_real`), `observation_stream_hz`, and
`validate_checkpoints` (`"true"` opts the executor into enforcing each
subgraph's `validate=True` postcondition checkpoints at real-execution
time — **currently a no-op**; reserved seam).

## SubgraphDef shape

```json
{
  "skill":  "<MORSL skill name; echoed from your spec>",
  "inputs":  { "<name>": "<proto.type.string>" },
  "outputs": { "<name>": { "$ref": "<node>.<field>" } },
  "nodes": {
    "<node_name>": { /* NodeDef */ },
    "<terminal_marker>": { "type": "noop" }
  },
  "edges": [
    ["START", "<first_node>"],
    ["<first_node>", "<second_node>"],
    ["<terminal_marker>", "END"]
  ],
  "conditional_edges": { },
  "exit": { "router_field": null, "success_values": ["<success_exit>"] },
  "on_error": "<failure_exit_value>"
}
```

Scripts go in separate fenced blocks, namespaced under
`scripts/<subgraph_name>/`:

````python:scripts/<subgraph_name>/<script>.py
...
````

## Node types

The two production dispatch types are `tool` and `script`. `noop` and
`router` are control-flow markers. `subgraph` and `end` only appear at
the top level of the workflow (not inside subgraphs).

```json
{ "type": "tool", "tool": "gripper.Open",
  "inputs": { "settle_steps": 40 } }                       /* gRPC method */

{ "type": "tool", "tool": "sam3.SegmentBox",
  "inputs": { "image": { "$ref": "obs.rgb" }, "box": { "$ref": "perceive.box" } } }

{ "type": "tool", "tool": "run_policy",
  "inputs": { "observation_stream": { "$ref": "in.observation_stream" } } }  /* atomic MORSL skill */

{ "type": "tool", "tool": "libero_pi05",
  "inputs": { "prompt": "...", "max_windows": 25 } }       /* learned policy */

{ "type": "tool", "tool": "track_object", "streaming": true,
  "inputs": { "observation_stream": { "$ref": "in.observation_stream" } } }

{ "type": "script", "script": "scripts/<subgraph_name>/foo.py",
  "inputs": { ... } }                                      /* canonical bundle script
                                                              or rare inline helper */

{ "type": "noop" }                                          /* named terminal marker */
{ "type": "router", "script": "scripts/<sg>/route.py", "inputs": { ... } }
```

- **`tool`** — the canonical dispatch for **anything**: gRPC service
  methods (auto-registered as `<package>.<MethodName>`, e.g.
  `observation.GetObservation`, `gripper.Open`,
  `geometry_svc.FilterAndComputeOBB`), MORSL skills (registered by skill
  name, e.g. `run_policy`, `track_object`), and learned policies
  (registered by policy name). `tool:` is always a flat name; never write
  `<package>.v1.<Service>` or include a separate `method:` field. The
  runtime resolves the proto FQN internally.
- **`script`** — local Python file in the workflow folder. Prefer to
  point at a canonical bundle script (listed in the chosen skill's
  "Canonical scripts" table); only emit your own inline Python for
  ad-hoc helpers that no canonical and no tool covers.
- **`noop`** — empty body. Used as a named subgraph terminal so the node
  name becomes the subgraph's exit value when `exit.router_field=null`.
- **`router`** — Send dispatch. Function returns either a string target
  for static routing or a list of `{"to": "...", "inputs": {...}}` dicts
  for dynamic fan-out.

*Legacy node types (`type: service`, `type: skill`, `type: policy`)
have been retired. The validator rejects them with a precise migration
message. Always emit `type: tool` with the flat name instead.*

`streaming: true` is only valid on `tool` and `script` nodes. A
streaming node has **no outgoing edges** — it's a pure source. Consumers
read its latest published value via `{"$ref": "<node>"}`. The chosen
tool's bundle contract must declare `contract.streaming: true`.

## Edge semantics

- **Static edges**: `["src", "dst"]` pairs.
- **Multiple outgoing edges from one node = parallel super-step.**
- **`conditional_edges`** dispatches on a router field. From a
  non-router source the field is on the source's output (set
  `router_field` to its name). From a router source set
  `router_field: null`.
- **`exit.router_field: null`** means the subgraph's exit value is the
  name of the terminal node (the one whose edge points to `END`). For
  this to work, declare your terminals as `noop` nodes named after the
  exit values (e.g. `"found": {"type": "noop"}` with edge
  `["found", "END"]`).
- **`exit.router_field: "<field>"`** reads the field on the terminal
  node's output as the exit value.
- **`on_error`** is an optional subgraph-level catch: when any node
  raises, the runtime emits this string as the subgraph's exit value
  (bypassing the terminal-node read). The `on_error` symbol is the
  subgraph's single failure exit. It is **never** a declared node and
  **never** a `conditional_edges` mapping target — failures surface
  only by raising.

## Reference syntax

| Form | Meaning |
|---|---|
| `{"$ref": "observe"}` | Full output of the `observe` node. |
| `{"$ref": "observe.cameras"}` | Nested field walk on the output. |
| `{"$ref": "tracker"}` | Snapshot of the latest published value of the streaming `tracker` node. |
| `{"$ref": "in.<name>"}` | Cross-subgraph input from your declared `inputs` schema. |

## Validation rules

Your subgraph fails validation if:

1. Any node referenced by an edge or conditional-edge mapping is not
   declared in `nodes` (and is not `START`/`END`).
2. Any non-`END` node is unreachable from `START`.
3. Any non-streaming, non-end node has no outgoing edge or
   conditional-edges entry.
4. A streaming node has any outgoing edge or conditional-edges entry.
5. A node with `streaming: true` invokes a skill whose contract has
   `streaming: false` (or vice versa) — when the skill registry is
   available.
6. `conditional_edges` from a non-router source omits `router_field`.
7. `conditional_edges` from a router source sets `router_field` to
   non-null (router scripts return the target directly).
8. `outputs` binding references an unknown node or an end node.
9. `exit.success_values` is empty (S7).
10. `on_error` collides with a declared node (S9) or appears as a
    `conditional_edges` mapping target (S10).
11. With `router_field: null`, any name in `exit.success_values` is
    not declared as a `noop` node; OR with `router_field` set, any
    name in `exit.success_values` collides with a node name (S11).
12. Any `{"$ref": "in.<name>"}` references an input name not declared
    in your `inputs` schema (the executor-injected
    `observation_stream` is exempt).
13. `inputs.<name>` declared on a reachable subgraph has no upstream
    producer subgraph that declares an output of the same name.

Errors are fed back; fix every error and re-emit the full subgraph JSON.

## v2 → v3 mapping (for migration)

| v2 concept | v3 replacement |
|---|---|
| `begin: <state_name>` | `["START", node]` edge |
| EndState marker (`{"type": "end"}` inside states) | `noop` node + edge to `END` |
| `on_success` | static edge to next node |
| `on_failure` | subgraph-level `on_error: "<exit>"` catch (or per-node `conditional_edges`) |
| `parallel` state with `branches` | multiple outgoing edges from one node |
| `join_policy: first_success` (race) | NOT in graph — long-running concurrent participants become `streaming: true` skill nodes (consumed via `$ref` snapshots) |
| `transitions: { exit_cond: target_sg }` | top-level `conditional_edges` on the subgraph node, `router_field: "exit"` |
| EndSubgraph | top-level `end`-type node with `status` and `recovery` |

## Type mapping

| Python | Proto |
|---|---|
| `Vec3`, `Se3Pose`, `Quaternion` | `common.Vec3`, etc. |
| `list[Se3Pose]` | `repeated common.Se3Pose` |
| `PointCloud \| None` | optional `common.PointCloud` |
| `float` | `double` |
| `int` | `int64` |

Proto imports — the two `common` modules split as follows.
**`OrientedBoundingBox` lives in `geometry_pb2`, not `sensor_pb2`** —
mis-importing it from `sensor_pb2` is the leading cause of script-load
`ImportError`s in this codebase:

```python
# vos.proto.common.geometry_pb2 — pure geometry types
from vos.proto.common.geometry_pb2 import (
    Vec3, Quaternion, Se3Pose, OrientedBoundingBox,
)

# vos.proto.common.sensor_pb2 — sensor / perception payload types
from vos.proto.common.sensor_pb2 import CameraObservation, Mask, PointCloud

# Per-service request/response messages
from vos.proto.observation.v1 import observation_pb2
from vos.proto.geometry_svc.v1 import geometry_svc_pb2
from vos.proto.curobo.v1 import curobo_pb2
from vos.proto.robot_control.v1 import robot_control_pb2
from vos.proto.gripper.v1 import gripper_pb2
```

## Proto field reference

Exact field names — these trip up LLMs:

| Message | Fields |
|---|---|
| `Vec3` | `x`, `y`, `z` (all `double`) |
| `Quaternion` | `w`, `x`, `y`, `z` (WXYZ convention). Top-down gripper is `Quaternion(w=0, x=1, y=0, z=0)`. |
| `Se3Pose` | `position: Vec3`, **`rotation: Quaternion`** (NOT `orientation`) |
| `OrientedBoundingBox` | `center: Vec3`, `extent: Vec3` (half-extents), **`orientation: Quaternion`** (NOT `rotation`) |

Note the asymmetry: `Se3Pose.rotation` vs `OrientedBoundingBox.orientation`.


## Skill in scope: `perception_single`

# perception_single

Single-path perception: detect → disambiguate → segment → fuse to 3D →
extract OBB. The full pipeline runs once per camera; results across cameras
are merged via KD-tree intersection inside `perceive_dino_vlm.py`.

## When to use

- Uncluttered scenes with visually distinct targets.
- Platforms where only DINO + VLM + SAM3 + Geometry are deployed.
- When `perception_multi` is not in the available skill catalog.

## When NOT to use

- Cluttered scenes with similar nearby distractors. Prefer `perception_multi`.

## Recommended subgraph state flow

3 states:

```text
observe → perceive → filter_obb
```

State details:

> **About `object_name` below:** it is a literal Python string — the natural
> noun phrase for the object you are perceiving, drawn from this subgraph's
> description (e.g. `"alphabet soup can"`, `"basket"`, `"red bowl"`). It is
> a constant per subgraph instance, NOT a binding. **DO NOT** write
> `Ref("in.object_name")` or any other `Ref(...)`; the coordinator does
> not declare `object_name` as a subgraph input. Write the string directly,
> e.g. `"object_name": "basket"`.

1. **`observe`** — `type: tool`, `tool: "observation.GetObservation"`,
   `inputs: {}`. Auto-registered gRPC method; flat name only.
2. **`perceive`** — `type: script`, file `scripts/<sg>/perceive_dino_vlm.py`
   from this bundle. Inputs:
   `cameras=Ref("observe.cameras")`,
   `object_name="basket"` (replace with the actual target noun phrase
   from this subgraph's description),
   plus any optional fields (`object_description`, `dino_prompt`, etc.).
   Returns `{found, cloud, mask, score}`.
3. **`filter_obb`** — `type: tool`,
   `tool: "geometry_svc.FilterAndComputeOBB"`,
   `inputs={"point_cloud": Ref("perceive.cloud")}`. Returns a bare
   `OrientedBoundingBox`.

### Wiring the exit (HARD)

Use the linear edge `filter_obb → found → END`. The `perceive` script
already raises if the target isn't found, so the subgraph's
`on_error: "not_found"` catches that path automatically. Do **NOT**
add any conditional edges on `perceive` — the linear path plus
`set_on_error` is sufficient.

✅ Correct (the literal `vos.builder` calls you should emit):

```python
sg.add_node("filter_obb", type="tool",
            tool="geometry_svc.FilterAndComputeOBB",
            inputs={"point_cloud": Ref("perceive.cloud")})

# add_exit() creates the success-marker noop node AND registers the
# exit value. Do NOT also call sg.add_node("found", type="noop") — that
# would conflict with the node add_exit created.
sg.add_exit("found")

sg.add_edge("perceive", "filter_obb")
sg.add_edge("filter_obb", "found")
sg.add_edge("found", END)

sg.set_on_error("not_found")
```

Bind the subgraph outputs (ALL THREE — required, no exceptions):

```python
sg.set_outputs(
    target_obb=Ref("filter_obb"),
    target_mask=Ref("perceive.mask"),
    target_cloud=Ref("perceive.cloud"),
)
```

(Replace `target_*` with this subgraph's actual name prefix — e.g.
`container_obb`, `container_mask`, `container_cloud` when authoring
the container subgraph.)

Note the asymmetry: `<name>_obb` references `filter_obb` with no
trailing field (FilterAndComputeOBB returns a bare OBB), while
`<name>_mask` and `<name>_cloud` walk into fields of `perceive`'s
output dict. `perceive_dino_vlm.py` already produces all three;
emitting them unconditionally lets downstream subgraphs that need any
of them (e.g. `grasp_moe` requires `<name>_cloud`) wire up without you
having to anticipate which skill they'll use.

## Hard rules

1. Subgraph-level outputs MUST emit ALL THREE: `<name>_obb`,
   `<name>_mask`, AND `<name>_cloud`. The cloud is the fused world-frame
   point cloud needed by learned-grasp skills (e.g. `grasp_moe`); emit
   it unconditionally so the downstream agent can wire it without
   round-tripping. See `references/perception_pipeline_invariants.md`.
2. `geometry_svc.FilterAndComputeOBB` returns a bare `OrientedBoundingBox`;
   bind via `Ref("filter_obb")` (no trailing field).
   See `references/geometry_calling_conventions.md`.

## Required end states

| End state | Meaning |
|---|---|
| `found` | OBB + mask bound; route to next subgraph (typically `grasp_*`). |
| `not_found` | Route to `abort` (or to `done` in clean-all-items loops). |


## See also

- `references/single_vs_multi.md` — the choice between single- and multi-
  method perception.
- `prompts/vlm_select_box.md` — the inline VLM prompt template.
- `scripts/perceive_dino_vlm.py` — the canonical perception script.

### Canonical scripts (you may emit as `type: script` states)

| Logical name | Path | Inputs | Outputs |
|--------------|------|--------|---------|
| `perceive_dino_vlm` | `scripts/perceive_dino_vlm.py` | cameras: list[CameraObservation], object_name: str, text_prompts: list[str] | None, min_points: int, min_score: float, use_multiview: bool, box_threshold: float, text_threshold: float, dino_prompt: str, object_description: str | found: bool, cloud: PointCloud, mask: Mask, score: float |

## Tools available in this subgraph

| Tool | Transport | Summary |
|------|-----------|---------|
| `geometry_svc.FilterAndComputeOBB` | grpc | Geometry.FilterAndComputeOBB :: FilterAndComputeOBBRequest → OrientedBoundingBox |
| `geometry_svc.MaskToWorldPoints` | grpc | Geometry.MaskToWorldPoints :: MaskToWorldRequest → PointCloud |
| `grounding_dino.Detect` | grpc | GroundingDino.Detect :: GroundingDinoDetectRequest → GroundingDinoDetectResponse |
| `observation.GetObservation` | grpc | Observation.GetObservation :: Empty → ObservationResponse |
| `sam3.SegmentBox` | grpc | Sam3.SegmentBox :: Sam3SegmentBoxRequest → SegmentResponse |
| `sam3.SegmentText` | grpc | Sam3.SegmentText :: Sam3SegmentTextRequest → SegmentResponse |
| `vlm.Query` | grpc | Vlm.Query :: VlmQueryRequest → VlmQueryResponse |

## Your subgraph: `container_sg`

Locate the target location

### Required outputs
Bind these from internal state fields via `{"$ref": "<state>.<field>"}` (or `{"$ref": "<state>"}` for bare-message returns):

| Name | Type |
|------|------|
| container_obb | `OrientedBoundingBox` |
| container_mask | `Mask` |
| container_cloud | `PointCloud` |

### Required end states
Your subgraph must contain exactly these end states. Each is a `{"type": "end"}` marker reached via some `on_success`/`on_failure` chain.

| End state | Meaning |
|-----------|---------|
| `found` | Target detected; OBB and mask bound in subgraph outputs. |
| `not_found` | Target not visible in any view. In clean-all-items loops route to done; in normal pick-and-place, route to abort. |

### Upstream outputs (already produced by earlier subgraphs)
| Subgraph | Output | Type |
|----------|--------|------|
| target_sg | target_obb | `OrientedBoundingBox` |
| target_sg | target_mask | `Mask` |
| target_sg | target_cloud | `PointCloud` |
\end{lstlisting}

\section{Sample Generation Outputs}

\begin{lstlisting}
    ```python
from vos.builder import Subgraph, Ref, START, END

sg = Subgraph(name="container_sg", skill="perception_single")

# Add nodes for the perception pipeline
sg.add_node("observe", type="tool", tool="observation.GetObservation")

sg.add_node("perceive", type="script", 
            script="scripts/perceive_dino_vlm.py",
            inputs={
                "cameras": Ref("observe.cameras"),
                "object_name": "container"
            })

sg.add_node("filter_obb", type="tool",
            tool="geometry_svc.FilterAndComputeOBB",
            inputs={"point_cloud": Ref("perceive.cloud")})

# Success marker
sg.add_exit("found")

# Wiring
sg.add_edge(START, "observe")
sg.add_edge("observe", "perceive")
sg.add_edge("perceive", "filter_obb")
sg.add_edge("filter_obb", "found")
sg.add_edge("found", END)

# Bind outputs
sg.set_outputs(
    container_obb=Ref("filter_obb"),
    container_mask=Ref("perceive.mask"),
    container_cloud=Ref("perceive.cloud")
)

# Failure path
sg.set_on_error("not_found")
```
\end{lstlisting}

\end{document}